\pdfoutput=1

\documentclass[11pt]{article}

\usepackage[]{acl}

\usepackage{times}
\usepackage{latexsym}

\usepackage[T1]{fontenc}

\usepackage[utf8]{inputenc}

\usepackage{microtype}

\usepackage{inconsolata}

\usepackage{times}
\usepackage{latexsym}

\usepackage{longtable}
\usepackage{tabu}
\usepackage{arydshln}

\usepackage{xspace}
\usepackage{graphicx}
\usepackage{wrapfig}
\usepackage{CJK,algorithm,algorithmic,amssymb,amsmath,array,epsfig,graphics,float,subfigure,verbatim,epstopdf}
\usepackage{enumitem}
\usepackage{color,soul}
\usepackage{hhline}
\usepackage{multirow}
\usepackage{multicol}
\usepackage{xcolor}
\usepackage{hyperref}
\usepackage{cleveref}

\usepackage{xurl}

\usepackage{tabularx}
\usepackage{bm}
\usepackage{seqsplit}
\usepackage{stmaryrd}
\usepackage{booktabs}

\usepackage{comment} 

\usepackage{xparse}

\NewDocumentCommand{\heng}
{ mO{} }{\textcolor{red}{\textsuperscript{\textit{heng}}\textsf{\textbf{\small[#1]}}}}

\NewDocumentCommand{\qingyun}
{ mO{} }{\textcolor{cyan}{\textsuperscript{\textit{qingyun}}\textsf{\textbf{\small[#1]}}}}

\NewDocumentCommand{\doug}
{ mO{} }{\textcolor{brown}{\textsuperscript{\textit{doug}}\textsf{\textbf{\small[#1]}}}}
\newcommand*\inlinelargeimage[1]{\raisebox{-0.15\baselineskip}{$\,$\includegraphics[height=0.9\baselineskip]{#1}$\,\,$}}

\newcommand{\testtube}{\inlinelargeimage{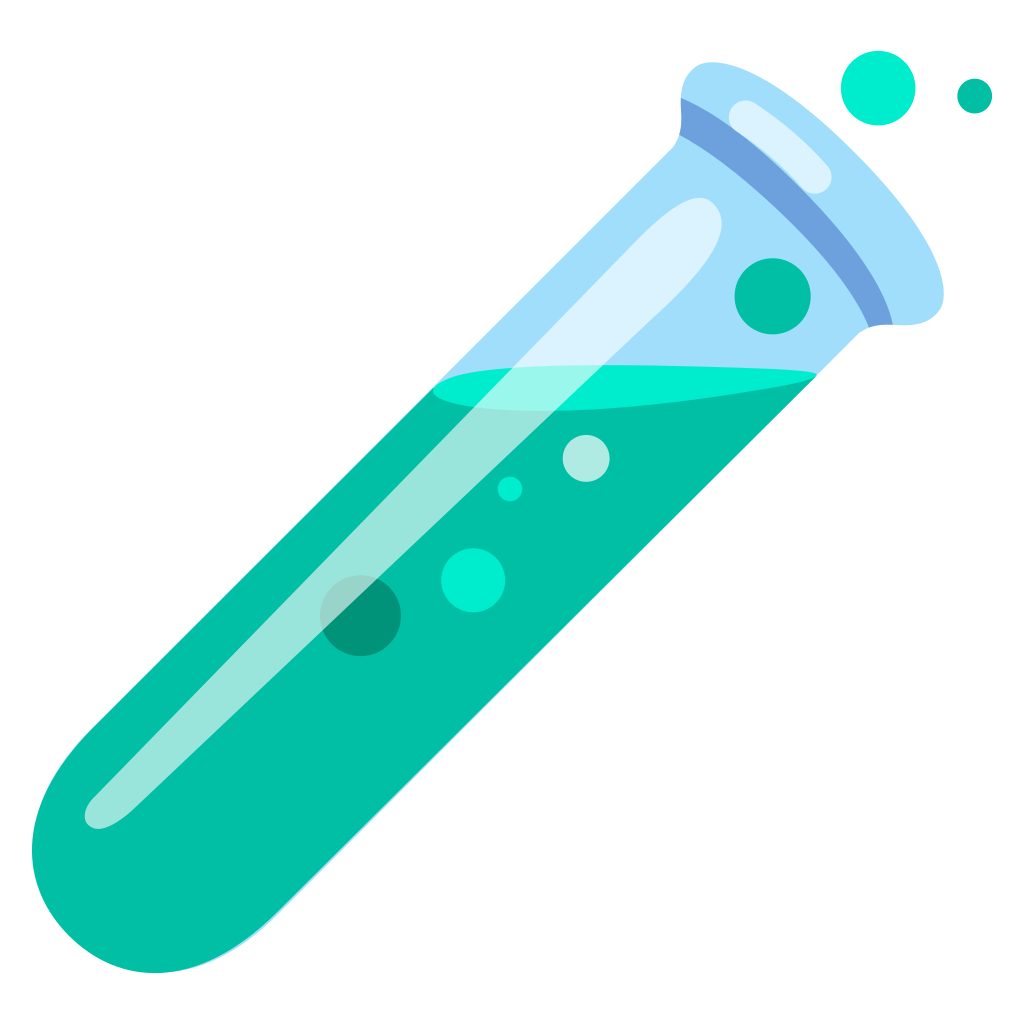}}

\title{\textsc{SciMON}\testtube: Scientific Inspiration Machines Optimized for Novelty}

\author{
Qingyun Wang$^{1}$, \  Doug Downey$^{2}$,  \textbf{Heng Ji}$^{1}$, \ Tom Hope$^{2,3}$ \\   
$^{1}$ University of Illinois at Urbana-Champaign $^{2}$ Allen Institute for Artificial Intelligence (AI2) \\  $^{3}$ The Hebrew University of Jerusalem  \\
\texttt{\fontfamily{pcr}\selectfont \{tomh,doug\}@allenai.org},
\texttt{\fontfamily{pcr}\selectfont\{qingyun4,hengji\}@illinois.edu}\\
}

\begin{document}
\maketitle

\begin{abstract}
We explore and enhance the ability of neural language models to generate novel scientific directions grounded in literature. Work on literature-based hypothesis generation has traditionally focused on binary link prediction---severely limiting the expressivity of hypotheses. This line of work also does not focus on optimizing novelty. 
We take a dramatic departure with a novel setting in which models use as input background contexts (e.g., problems, experimental settings, goals), and output \emph{natural language ideas} grounded in literature. We present \textsc{SciMON}, a modeling framework that uses retrieval of ``inspirations'' from past scientific papers, and explicitly optimizes for novelty by iteratively comparing to prior papers and updating idea suggestions until sufficient novelty is achieved.    
Comprehensive evaluations reveal that GPT-4 tends to generate ideas with overall low technical depth and novelty, while our methods partially mitigate this issue. Our work represents a first step toward evaluating and developing language models that generate new ideas derived from the scientific literature\footnote{Code, data, and resources are publicly available for research purposes: \url{https://github.com/eaglew/clbd}.}.

\end{abstract}

\section{Introduction}
Can machines mine scientific papers and learn to suggest new directions? The idea that information from the literature can be used for automatically generating hypotheses has been around for decades \cite{swanson1986undiscovered}. To date, the focus has been on a specific setting: hypothesizing links between pairs of concepts (often in drug discovery applications \cite{henry2017literature}, e.g., new drug-disease links), where concepts are obtained from papers or knowledge bases previously derived from papers \cite{sybrandt2020agatha, nadkarni2021scientific}.

\begin{figure}
\centering
\includegraphics[width=0.98\linewidth]{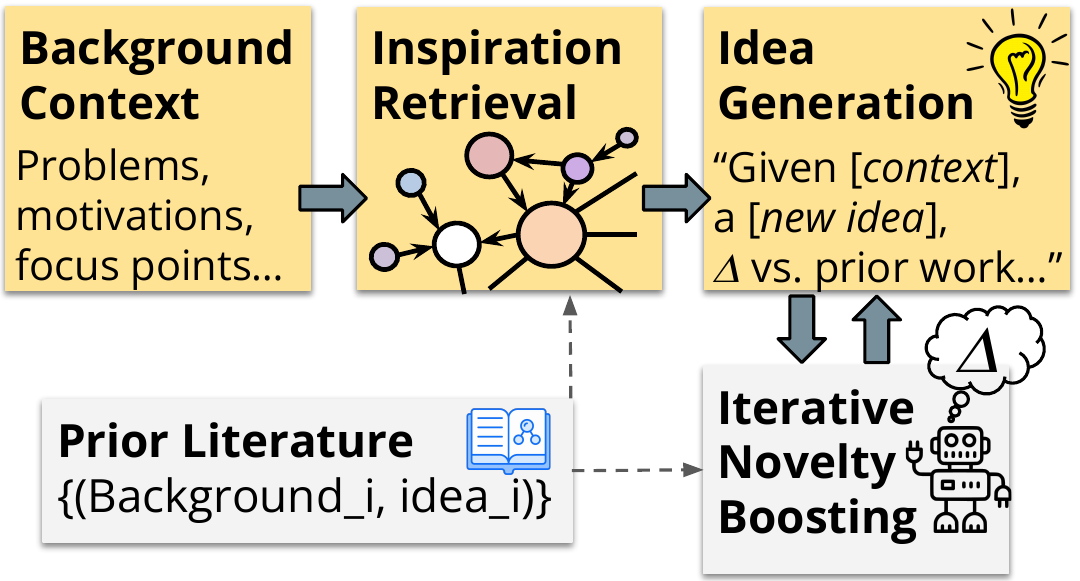}
\caption{\label{fig:overview} \textsc{SciMON} takes background context and generates ideas grounded in literature inspirations, optimizing novelty by iteratively comparing to related work.
}
\end{figure}

This common setting has fundamental drawbacks. Reducing the ``language of scientific ideas'' \cite{hope2022computational} to this simplistic form limits the expressivity of the hypotheses we can hope to generate, and does not capture nuanced \emph{contexts} that scientists consider: target application settings, requirements and constraints, motivations and challenges. In light of the strong progress recently made with large language models (LLMs), in this paper we explore a dramatically different setting: models that take descriptions of problem contexts---and return \emph{natural language} suggestions of {novel} scientific directions that are grounded in literature.

We develop a framework named \textsc{SciMON} (Scientific Inspiration Machines with Optimization for Novelty), named after Nobel laureate and AI pioneer Herbert Simon who authored early foundational work on automated scientific discovery \cite{newell1956logic,simon1973does}. We first present an automated data collection methodology that collects examples of past problems and proposed ideas from scientific papers. We then use this data for both fine-tuning and in-context training of LLMs---training them to take problem descriptions and output proposed ideas to address them. We observe that state-of-art LLMs (e.g., GPT-4 \cite{openai2023gpt4}) struggle with generating novel scientific ideas, and contribute a new modeling framework for generating  hypotheses that makes progress in improving the hypothesis generation ability of LLMs (Figure \ref{fig:overview}). Given a background problem description, models first dynamically retrieve \emph{inspirations} from past literature in the form of related problems and their solutions along with contexts from a scientific knowledge graph. These retrieved inspirations serve to ground the generated ideas in existing literature. We then endow models with the ability to iteratively boost the \emph{novelty} of generated ideas. Given an idea $\mathcal{I}$ generated by the LLM at step $t$, the model compares $\mathcal{I}$ with existing research in the literature; if it finds strongly overlapping research, the model is tasked with updating its idea to be more novel relative to prior work (much like a good researcher would do). We also introduce an \emph{in-context contrastive model} which encourages novelty with respect to background context.

We perform the first comprehensive evaluation of language models for generating scientific ideas in our new generative, contextual setting. We focus on AI/NLP ideas to facilitate analysis by AI researchers themselves,
and also demonstrate generalization to the biomedical domain. We design extensive evaluation experiments using human annotators with domain expertise to assess relevance, utility, novelty, and technical depth. 
Our methods substantially improve the ability of LLMs in our task; however, analyses show that ideas still fall far behind scientific papers in terms of novelty, depth and utility---raising fundamental challenges toward building models that generate scientific ideas.

\section{Background and New Setting}
\label{sec:task}

\begin{figure*}[t]
\centering
\includegraphics[width=\linewidth]{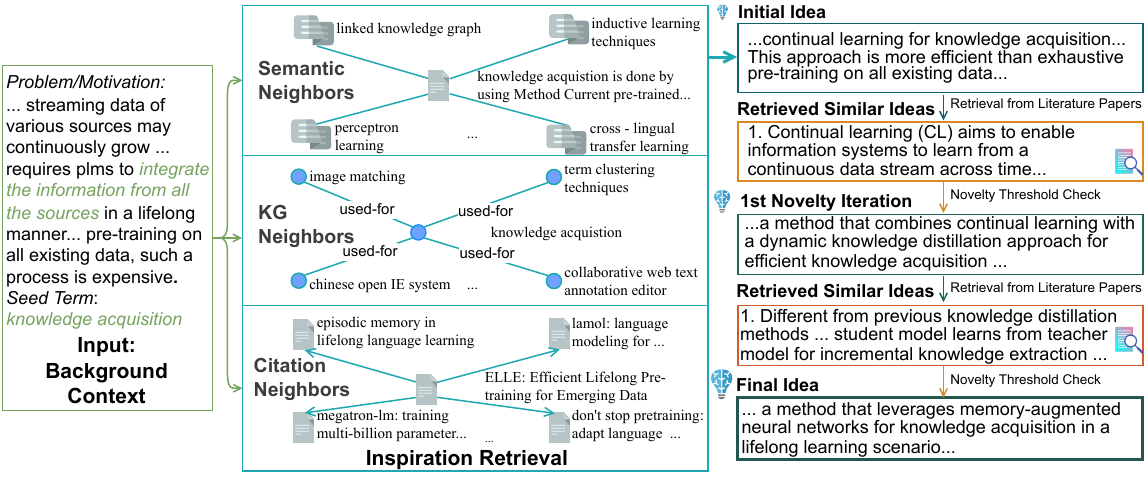}
\caption{\label{fig:retrieval} Architecture overview. Our models retrieve \emph{inspirations} and then pass the background input and retrieved inspirations to an LM-based generation module, which iteratively optimizes novelty. 
Input from \citet{qin-etal-2022-elle}.
}
\end{figure*}

We begin with a brief description of related work and background. We then present our novel setting. 

\paragraph{Literature-based discovery} Nearly four decades have passed since Don Swanson pioneered Literature-Based Discovery (LBD), based on the premise that the literature can be used for generating hypotheses \cite{swanson1986undiscovered}. LBD has been focused on a very specific, narrow type of hypothesis: links between pairs of concepts (often drugs/diseases). The classic formalization of LBD goes back to  \citet{swanson1986undiscovered} who proposed the ``ABC'' model where two concepts (terms) A and C are hypothesized as linked if they both co-occur with some intermediate concept B in papers. 
More recent work has used word vectors~\cite{tshitoyan2019unsupervised} or link prediction models \cite{wang-etal-2019-paperrobot,sybrandt2020agatha,xu-etal-2023-exploring} to discover scientific hypotheses as pairwise links between concepts. A tightly related body of research focuses on \emph{scientific knowledge graph link prediction} \cite{nadkarni2021scientific}, where predicted links may correspond to new hypotheses, and knowledge bases are reflections of existing scientific knowledge in specific domains, derived from literature. A fundamental gap in this line of work is in the lack of approaches for modeling nuanced contexts \cite{sosa2022contexts} (e.g., the {specific} settings in which a drug may be relevant for a disease) for generating ideas in open-ended problem settings with unbounded hypothesis spaces, and for optimizing novelty.
Our setting can be viewed as a radical departure addressing the limitations in existing settings. 

\paragraph{LLMs for Scientific Innovation} Large language models (LLMs) have made remarkable progress in interpreting and producing natural language content and handling knowledge-intensive tasks such as in the medical domain \cite{nori2023capabilities}. Very recent work \cite{boiko2023autonomous} has explored the use of LLMs in a robotic chemistry lab setting, planning chemical syntheses of known compounds and executing experiments. Robotic lab settings are inherently limited to narrow sub-areas where such experiments are possible and relevant. 
Other very recent work \cite{huang2023benchmarking} used LLMs to produce code for machine learning tasks such as Kaggle competitions, finding that a GPT-4 agent achieved 0\% accuracy on research challenges such as BabyLM \cite{warstadt2023findings}. GPT-4 has been anecdotally reported as having ``strengths less like those of having a human co-author, and more like a mathematician working with a calculator'' \cite{carlini2023llm}. 
Our goal is to conduct a non-anecdotal evaluation and enhancement of strong LLMs' ability to generate novel open-ended scientific ideas.

\subsection{\textsc{SciMON} Problem Setting}
\label{sec:prb}

We are motivated by imagining an AI-based assistant that suggests ideas in natural language. The assistant takes as input background context $\mathcal{B}$ consisting of (1) current problems, motivations, experimental settings and constraints, denoted as $\mathcal{M}$; and optionally (2) a seed term $v$ that should be a focus point of the generated idea $\mathcal{I}$. The seed term is motivated by considering a user-provided cue for the model to limit its hypothesis space. Importantly, generated ideas should not merely paraphrase the background---the output should be \emph{novel} with respect to $\mathcal{B}$ and the broader literature corpus. Figure~\ref{fig:retrieval} illustrates the setting, showing a background text that describes problems with \textit{``pretrained language models''} in the lifelong integration of information sources, including computational costs. The assistant aims to generate an idea for performing \textit{``knowledge acquisition''} within this context. Given this input, we aim to generate a full sentence describing a novel idea. 

\begin{figure}
\centering
\includegraphics[width=0.98\linewidth]{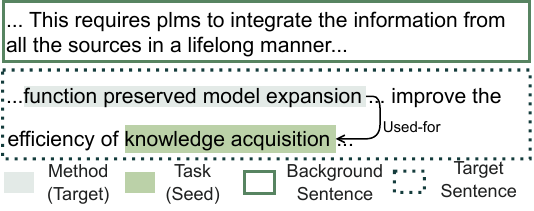}
\caption{\label{fig:small-ie} 
We use IE to obtain literature data for our approach: problems/motivations (background) and proposed ideas (target), as well as salient seed terms.
}
\end{figure}

\subsection{Automated Training Data Collection}
\label{sec:datadetail}

We obtain training data derived from papers with scientific information extraction (IE) models---extracting past examples of background sentences and corresponding ideas (e.g., descriptions of methods used for specific problems in the background sentences), along with salient entities as seed terms. This data is used for training in both in-context learning and fine-tuning setups.

We construct a corpus $\mathcal{D}$ from 67,408 ACL Anthology papers from S2ORC~\cite{lo-etal-2020-s2orc} (we later also conduct an experiment with a biomedical corpus \S\ref{sec:human}). Given a title and the corresponding abstract from a document $d$, to select problem/motivation sentences $\mathcal{M}$ we first perform scientific sentence classification~\cite{cohan-etal-2019-pretrained} to classify sentences from the abstract into one of \{\textit{Background}, \textit{Method}, \textit{Objective}\}, selecting sentences with labels of \textit{Background} and treating the remaining sentences as target sentences $\mathcal{T}$ which will serve as desired output examples (Figure \ref{fig:small-ie}). 

For seed term selection, we apply a state-of-the-art scientific IE system~\cite{ye-etal-2022-packed} to  $\mathcal{T}$ to extract \emph{entities} corresponding to \textit{Task}, \textit{Method}, \textit{Evaluation Metric}, \textit{Material}, and \emph{relations}
of the form \texttt{[method,used-for,task]}---mentions of methods and the tasks they are used for, materials used for tasks, etc. We treat the head (e.g., method) or tail (e.g., task) entity as the \emph{seed} term, and name the other entity (tail/head, respectively) as a \textit{target} term $t \in \mathcal{T}$.
Continuing our example from Figure \ref{fig:retrieval}, Figure \ref{fig:small-ie} shows how the seed and target terms (``\textit{knowledge acquisition}'' and ``\textit{function preserved model expansion}'') are extracted from $\mathcal{T}$.
During training, each instance contains ($\mathcal{B}$,$\mathcal{T}$) pairs; during evaluation, target information is removed.

We use SciCo~\cite{cattan2021scico} to obtain coreference links for entity normalization, and use ScispaCy~\cite{neumann-etal-2019-scispacy} to replace abbreviations with a more informative long form. We also collect paper metadata, including the citation network $\mathcal{G}_c$. We split our dataset temporally (train/dev/test correspond to papers from years $<$2021 / 2021 / 2022 respectively). For our experiments, we used model checkpoints trained on data preceding 2022, avoiding the risk of data contamination (§\ref{sec:limitation}). Table \ref{tab:stat} shows data statistics.\footnote{More details are in Appendix~\ref{app:he_ap}.}

\paragraph{Quality of IE Preprocessing} During preprocessing, we only keep high-confidence outputs from IE models to reduce errors. We observe this removes many of the noisy cases. To validate this, we manually evaluate the precision of each preprocessing step on a random sample of papers and observe that all steps yield high precision (91\%-100\%) except relation extraction (65\%); in total, the rate of instances passing all steps was 79.7\%.\footnote{See Table~\ref{tab:hum_pre} in Appendix.} 

\paragraph{Gold Test Set} We create a high-quality, clean test set. We remove test instances where models can trivially use surface-level background information to infer the ground truth to create a more \textit{challenging set}, selecting instances with low similarity between background and ground truth sentences. We compute the cosine similarity between each instance's background and corresponding ground truth sentence in the test set and take pairs with similarity $\leq0.074$, which amounts to the tenth percentile of pairs. We further annotate this subset to create a \textit{gold subset}. We manually exclude instances with trivial overlap between ground truth and background, remove cases with irrelevant background, and retain only instances where the target relation (from which the seed term is taken) is salient to the target sentence. We also remove test pairs that have unexplained terms in the background. We obtain a total of 194 instances.\footnote{Full annotation details are in Appendix~\ref{app:he_ap}.}

\begin{table}[!htb]
\centering
\small
\begin{tabularx}{\linewidth}{>{\arraybackslash\hsize=1\hsize}X>{\arraybackslash\hsize=1\hsize}X>{\arraybackslash\hsize=1\hsize}X>{\arraybackslash\hsize=1\hsize}X}
\toprule
\textbf{Split}  & \textbf{Forward } & \textbf{Backward} & \textbf{Total}\\ 
\midrule
Train      & 55,884 & 58,426 & 114,310  \\
Valid      & 7,938  & 8,257  & 16,195  \\
Test       & 2,623  & 2,686  &  5,309  \\
\bottomrule
\end{tabularx}
\caption{Dataset statistics. \label{tab:stat}Considering a relation of the form \texttt{[$v$ used-for $u$]}, we define \texttt{[$v$ used-for ?]} as \textit{forward}, and \texttt{[? used-for $u$]} as \textit{backward}.
}
\end{table}

\section{\textsc{SciMON} Models}
We present a new module to retrieve inspirations as contextual input (\S\ref{sec:retrieval}). Then, we describe another module to generate ideas given the context+inspiration (\S\ref{sec:baseline}). 
Finally, we introduce a new iterative novelty optimization method to further improve idea quality (\S\ref{sec:regeneration}).\footnote{ Training and hyperparameter details in Appendix \ref{sec:ft}.} 

\subsection{Inspiration Retrieval Module}
\label{sec:retrieval}

We take broad inspiration from cognitive aspects of innovation \cite{hope2022computational}: when researchers generate a new idea, they are grounded in a web of existing concepts and papers bearing on the new idea. We aim to enrich the context of each background by retrieving ``inspirations''--- pieces of information that can guide hypothesis generation. As illustrated in Figure~\ref{fig:retrieval}, for a given instance of the \textsc{SciMON} task, our retrieval augmentation can retrieve from three types of sources. Each source uses a different form of query and output.

\paragraph{Semantic Neighbors} 
For a given problem/motivation as input, ideas proposed for related problems in the training set can serve as a guiding reference for generating a new idea.
Given the background context $\mathcal{B}$ with a seed term $v$ and problem/motivation $\mathcal{M}$, we construct a base input $b$: a concatenation of $\mathcal{M}$ with a prompt $\mathcal{P}$ belonging to one of two templates: ``$v$ is used for $p$'' or ``$v$ is done by using $p$'', where $p$ is one of \textit{Task}/\textit{Method}/\textit{Material}/\textit{Metric}. In short, $b:=$ \textit{$\mathcal{P}$ $\oplus$ context:$\mathcal{M}$}. 
For example, in Figure \ref{fig:retrieval}, the concatenation is ``\textit{Knowledge acquisition is done by using Method; Context:...requires plms to integrate information...lifelong manner...}''.

We then retrieve inputs from the training set that are semantically related to a new base input $b$, and obtain target sentences $T$ corresponding to each retrieved training input. We extract the target term $t\in\mathcal{T}$ matching the seed term in $b$ (\S\ref{sec:datadetail}) as inspiration for input $b$. Simply put, this means we use as inspiration the salient aspect of the solution proposed in $\mathcal{T}$, which we found empirically to help remove noisy/irrelevant information in $\mathcal{T}$.
For example, in Figure~\ref{fig:retrieval}, we find
\textit{``informative entities are done by using Method context: in this work, we aim at equipping pre-trained language models with structured knowledge.''} 
as similar to the input and use $t=$\textit{``linked knowledge graph''} as inspiration.
 
Technically, we first construct a fully connected graph $\mathcal{G}_S$ based on the training set where each node is a pair of input text $b_i$ and target term $t_i$. We define the weight between two nodes $i$ and $j$ as the cosine similarity between $b_i$ and $b_j$ based on representations from SentenceBERT~\citep{reimers-gurevych-2019-sentence} (\texttt{all-mpnet-base-v2}). Given $b$, we first insert it into $\mathcal{G}_S$ and compute the weights of its connected edges. We then retrieve neighbors input text $\{b_1,\ldots,b_k\}$ from the training set with the largest edge weight, where $k$ is the number of retrieved instances. We consider the corresponding target terms $\{t_1,\ldots,t_k\}$ as semantic inspirations.

\paragraph{KG Neighbors}
We also explore enriching the context by linking it to a background KG with information on related methods and tasks. Using the same IE process used to extract our training examples (\S\ref{sec:datadetail}), we create a \emph{global} background KG $\mathcal{G}_B$ which covers all papers in the corpus $\mathcal{D}_\mathcal{Y}$ prior to a given year $\mathcal{Y}$ (i.e., the nodes in $\mathcal{G}_B$ correspond to tasks/methods/materials/metrics, and the edges are used-for relations, extracted and normalized from across the entire corpus as described earlier). Then, given a seed term $v$ at query time, we select adjacent nodes $\{n_1,n_2,...\}$ from $\mathcal{G}_B$ as inspirations. 
As an example, in Figure~\ref{fig:retrieval}, the neighbor nodes of \textit{``knowledge acquisition''} include \textit{``collaborative web text annotation editor''}, \textit{``image matching''}, etc., which we select as inspirations. 

\paragraph{Citation Neighbors} 
Another notion of contextual relatedness we explore is via citation graph links. Here, given as input background context $\mathcal{B}$, we assume access to the original source document $d$ from which $\mathcal{B}$ was extracted, and consider its \emph{cited} paper title set $\mathcal{C}_d$ as potential candidates. 
This can be seen as a stronger assumption on information available to the model--- assuming a researcher using the model provides relevant candidate documents from which ideas could be pooled. Because the training set only contains papers before year $\mathcal{Y}$, we only select papers $\mathcal{C}_{d\mathcal{Y}}\subseteq \mathcal{C}_d$ prior to year $\mathcal{Y}$. 
We then retrieve the top-$k$ titles with the highest cosine similarity to $d$ from $\mathcal{C}_{d\mathcal{Y}}$
based on their SentenceBERT embeddings as earlier. 
For instance, in Figure~\ref{fig:retrieval}, the paper ELLE~\cite{qin-etal-2022-elle} cites the paper \cite{NEURIPS2019_f8d2e80c}. Therefore, we choose the title \textit{``episodic memory in lifelong language learning''} as inspiration information.
\subsection{Generation Module}
\label{sec:baseline}
The idea generation module is given retrieved inspirations $i_1,\ldots,i_k$ along with context $\mathcal{M}$ as input.

\paragraph{In-Context Learning}
We experiment with recent state-of-the-art LLMs, GPT3.5 \texttt{davinci-003}~\cite{ouyang2022training} and GPT4 \texttt{gpt-4-0314} checkpoint~\cite{openai2023gpt4}. We first ask the model to generate sentences based on the seed term and the context in the zero-shot setting without any in-context examples (\texttt{GPT3.5ZS}, \texttt{GPT4ZS}). We then ask the model to generate sentences in a few-shot setting by prompting randomly chosen pairs of input and output from the training set (\texttt{GPT3.5FS}, \texttt{GPT4FS}).
Inspired by \citet{liu-etal-2022-makes}, we further employ a few-shot setting using semantically similar examples. Instead of random in-context examples, we use the top-$k$ examples from the training set with the highest cosine similarity to the query (\texttt{GPT3.5Retr}). 
This few-shot retrieval setting differs from the semantic neighbor discussed above, in that we provide both the input and output of each instance rather than solely supplying target entities as additional input.

\paragraph{Fine Tuning}
We fine-tune T5~\cite{JMLR:v21:20-074} (more recent models may be used too; see our biomedical experiment §\ref{sec:human} fine-tuning an LLM). 
We observe that the generation models tend to copy phrases from the background context. For example, given the context \textit{``...hierarchical tables challenge numerical reasoning ...''}, the model will generate \textit{``hierarchical table reasoning for question answering''} as the top prediction. For generating suggestions of \emph{novel} ideas, we wish to discourage overly copying from the background context. We introduce a new in-context contrastive objective, where negative examples are taken from the text in the input (e.g., in Figure~\ref{fig:retrieval}, the in-context negatives are \textit{plms}, \textit{pre-training}, etc). 
We compute an InfoNCE loss~\cite{oord2018representation} over the hidden states of the decoder, aiming to maximize the probability of the ground truth against those of in-context negatives:
\begin{align}
\begin{split}
    y^+&=\sigma(\mathrm{Avg}(\mathbf{W}_y\mathbf{h}^++\mathbf{b}_y))\\
    y^-_k&=\sigma(\mathrm{Avg}(\mathbf{W}_y\mathbf{h}^-_k+\mathbf{b}_y))\\
    \mathcal{L_\mathrm{cl}} &= \frac{\exp{\left(y^+/\tau\right)}}{\sum_k \exp{\left(y^-_k/\tau\right)} +\exp{\left(y^+/\tau\right)} } \\
\end{split}
\end{align}
where $\mathbf{h}^+$ and $\mathbf{h}_k^-$ are decoder hidden states from the positive and $k$-th negative samples, $\mathbf{W}_y$ and $\mathbf{b}_y$ are learnable parameters, $\sigma$ is a sigmoid function, $\tau$ is a temperature hyperparameter, and $\mathrm{Avg}(*)$ denotes the average pooling function based on the target sequence length. We optimize with both contrastive loss $\mathcal{L}_{\mathrm{cl}}$ and the cross-entropy loss. 

\subsection{Iterative Novelty Boosting with Retrieval}
\label{sec:regeneration}

We further improve the novelty of generated ideas with a new iterative \emph{retrieve-compare-update} scheme. Conceptually, we consider a novelty-inducing penalty $\gamma_\text{nov}(\mathcal{I}, \mathcal{R})$ that penalizes ideas $\mathcal{I}$ that are too ``close'' to existing ideas in literature reference examples $\mathcal{R}$. $\gamma_\text{nov}(\mathcal{I},\mathcal{R})$ is included during in-context learning and inference, providing numerical feedback in the form of a score reflecting similarity to existing work. We wish to minimize this score while ensuring $\mathcal{I}$ remains relevant to the background context $\mathcal{B}$; we do so iteratively by (1) retrieving related work from $\mathcal{R}$, (2) measuring degree of novelty, (3) instructing the model to update $\mathcal{I}$ to be more novel w.r.t $\mathcal{R}$, conditioning on $\mathcal{B}$. 

Specifically, in our implementation, we construct a reference corpus $\mathcal{R}$ based on all papers in the training set. We then propose an iterative algorithm that compares generated ideas against $\mathcal{R}$. We start with the initial idea $\mathcal{I}_0$ generated by the generation module. 
At each time step $t$, we use the generated idea $\mathcal{I}_t$ as a query to retrieve $k$ nearest ideas from the literature reference corpus $\mathcal{R}=\{R_1,...,R_k\}$ based on SentenceBERT, with the top-$k$ highest cosine similarity scores to $\mathcal{I}_t$ (we use $k=20$). For each retrieved ground truth literature idea $R_i$, we compare its cosine similarity score $S_i$ against a threshold $\mu$ (we use $0.6$). We provide all the retrieved ground truth ideas $\mathcal{\hat{R}}$ that pass the threshold as additional negative examples for the large language models with the following instruction prompt: ``\textit{Your idea has similarities with existing research as demonstrated by these} $j$ \textit{sentences: }$\mathcal{\hat{R}}$\textit{ Make sure the idea you suggest is significantly different from the existing research mentioned in the above sentences. Let's give it another try.}'' We stop the iteration once all $S_i$ are lower than $\mu$. Figure \ref{fig:retrieval} and Table \ref{tab:GPT4} demonstrate novelty iterations.

\section{Experiments}
\label{sec:exp}
\subsection{Human Evaluation}
\label{sec:human}
We present four human evaluation studies, exploring different facets of our problem and approach.

\subsubsection{Study I: Comparing Outputs across Model Variants}
We recruit six volunteer NLP experts with graduate-level education to rate the system. Raters are told to envision an AI assistant that suggests new paper ideas. We randomly select 50 instances (background+seed) from the gold subset. 
Each annotator receives ten instances, each paired with system outputs from different model variants (Table \ref{tab:human_evaluation}).
We ask raters to assess idea quality by considering each output's relevance to the context, novelty, clarity, and whether the idea is reasonable (positive ratings are dubbed “helpful” as shorthand, indicating they pass the multiple considerations).  
We observe moderately high rater agreement.\footnote{The agreement scores are in Table \ref{tab:alpha} Appendix \ref{app:he_ap}.} Raters are blind to the condition, and system outputs are randomly shuffled across instances. 

We instruct annotators to only provide positive ratings to ideas sufficiently different from the input context. In Study I, we ask raters not to anticipate groundbreaking novelty from the system but rather a narrower expectation of quality and utility; in Study II below, we enrich the analysis to examine \emph{ranking} between top models and also ``raise the bar'' and compare to actual ideas from papers.\footnote{Full evaluator guidelines are in Appendix \ref{app:he_ap}. The sample annotations are in Table~\ref{tab:human_anno}.} 

In a preliminary experiment, we also collected human ratings for \texttt{GPT4-ZS} (zero-shot) vs. \texttt{GPT4-FS} (few-shot) using the same criteria, finding \texttt{GPT4-FS} ranked higher in 65\% of cases, with the rest mostly tied; thus, zero-shot GPT-4 was left out of the remainder of study I and subsequent studies to reduce annotation effort and cost.

\paragraph{Results}
Overall, \texttt{GPT4FS} and \texttt{GPT4FS+KG} outperform other models by a wide margin (Table~\ref{tab:human_evaluation}). Apart from GPT4, \texttt{T5+SN+CL} performs best compared to other baselines, given its stronger prior knowledge of useful similar background hypotheses. In general, GPT3.5 models performed worse than fine-tuned T5 and its variants, which echoes results in other work in the scientific NLP domain \cite{jimenez-gutierrez-etal-2022-thinking}. GPT4 outputs tended to be longer, which may partially explain higher human preference. 

\begin{table}[!htb]
\centering
\small
\begin{tabularx}{\linewidth}{>{\hsize=0.8\hsize}X>{\centering\arraybackslash\hsize=0.8\hsize}X>{\centering\arraybackslash\hsize=0.8\hsize}X>{\centering\arraybackslash\hsize=1.4\hsize}X>{\centering\arraybackslash\hsize=1.4\hsize}X>{\centering\arraybackslash\hsize=0.7\hsize}X>{\centering\arraybackslash\hsize=1\hsize}X>{\centering\arraybackslash\hsize=0.7\hsize}X>{\centering\arraybackslash\hsize=1.4\hsize}X}
\toprule
\textbf{Type} &\texttt{3FS}& \texttt{3Rt}& \texttt{3FS+CT}& \texttt{3FS+KG}& \texttt{4}& \texttt{4+KG}& \texttt{T5}& {\texttt{T5+SN}}  \\ 
\midrule
H  &   33 & 25  & 16 & 33 & {73} & 66 & 22 &48\\
U  &   67 & 75  & 84 & 67 & 27 & 34 & 78 &52\\
\bottomrule
\end{tabularx}
\caption{Percent (\%) of total votes each system output receives from human raters. \textit{H} denotes a helpful output, while \textit{U} denotes an unhelpful output. ``\texttt{3FS}'' refers to the \texttt{GPT3.5FS}. ``\texttt{3Rt}'' refers to the \texttt{GPT3.5Retr}.  ``\texttt{4}'' refers to \texttt{GPT4FS}, and ``\texttt{4+KG}''  refers to the \texttt{GPT4FS+KG}. ``\texttt{T5+SN}'' refers to the \texttt{T5+SN+CL}. \texttt{GPT4FS} and \texttt{GPT4FS+KG} are rated much higher. While  \texttt{GPT4FS} has a slightly higher rating than the KG variant, a further human study reveals that \texttt{GPT4FS+KG} often leads to more technical depth (\S\ref{sec:human}).
\label{tab:human_evaluation}
}
\end{table}

\subsubsection{Study II: Comparing GPT4 Variants against Real Papers}
We conduct a follow-up human study of close competitors \texttt{GPT4FS} and \texttt{GPT4FS+KG} with a subset of the annotators to evaluate the incrementality and novelty of the generated ideas. In this study, model outputs are now \emph{ranked}, unlike the binary classification of helpful/not in Study I. Suggestions are ranked according to the level of technical detail and innovation in comparison to each other---i.e., ranking which of \texttt{GPT4FS} and \texttt{GPT4FS+KG} had a higher degree of technical detail and novelty, or whether they are roughly the same (tied). Finally, outputs are rated versus the ground truth idea, according to whether or not the suggestions were roughly at the same level of technical detail and innovation as the original paper's idea, or \emph{significantly lower}.

\paragraph{Results} 
Overall, \texttt{GPT4FS+KG} is found to have higher technical detail in 48\% of the compared pairs, and found to be less incremental (more novel) in 45\% of the pairs. Among the remaining 52\%/55\% (respectively), the vast majority are ties, indicating that whenever \texttt{GPT4FS+KG} is not favored, it is of roughly the same quality as  \texttt{GPT4FS}, but not vice versa. However, the most crucial aspect is comparing the results against the original ground truth idea on the quality of innovation. Here, we find that in 85\% of comparisons, the ground truth is considered to have \emph{significantly higher} technical level and novelty; and in the remaining 15\%, the ground truth was ambiguous or lacking additional context from the paper abstract. This points to a major challenge in obtaining high-quality idea generations using existing state-of-the-art models.

\subsubsection{Study III: Evaluation on Iterative Novelty Boosting}
We conduct a fine-grained evaluation of our novelty mechanism with qualitative and quantitative evaluation of novelty. Specifically, we ask five annotators to further compare the novelty-enhanced results against the initially generated ideas.  We randomly select 70 instances (background+seed) from the sentence generation gold subset. We ask annotators to check whether the new ideas are \emph{different} than the initial ideas (e.g., adding new information or approaches), and whether they are more \emph{novel} (i.e., a new idea can be different, but not necessarily more novel).
Since \texttt{GPT4FS+SN} outperforms other models, for this model, we further instruct annotators to compare the novelty of the second iteration results against the first iteration results. 

\begin{table}[!htb]
\centering
\small
\begin{tabularx}{\linewidth}{>{\hsize=2.2\hsize}X>{\centering\arraybackslash\hsize=1\hsize}X>{\centering\arraybackslash\hsize=0.6\hsize}X>{\centering\arraybackslash\hsize=0.6\hsize}X>{\centering\arraybackslash\hsize=0.6\hsize}X}
\toprule
\textbf{Type} &\texttt{GPT4FS}& \texttt{+SN}& \texttt{+CT}& \texttt{+KG}  \\ 
\midrule
1st Novelty $\Delta$ (\%)    &  +54.4  & +55.6  & +47.8  & +46.7 \\
2nd Novelty $\Delta$(\%)    & -  & +57.8  & - &  -\\\hdashline
1st new terms $\Delta$&  +23.1 & +22.8 & +22.1 & +21.9\\
2nd new terms $\Delta$&  -  & +21.5 & -  & - \\
\bottomrule
\end{tabularx}
\caption{Relative improvements of iterative novelty boosting. Iterations are applied to the ideas for which sufficiently similar related work is detected (\S\ref{sec:regeneration}).
``\texttt{1st Novelty}'' is \% of the 1st iteration ideas that gained \emph{novelty} over the initial idea, and ``\texttt{2nd Novelty}'' is the \% of gain over the 1st iteration. Our method substantially increases novelty for ideas to which it is applied. 
To save annotation resources, we only annotate second iteration results for the best-performing method (SN). We report the average number of new terms added, after filtering.
\label{tab:regeneration_evaluation}
} 
\end{table} 

\paragraph{Results}
For SN, in the first iteration 88.9\% of updated ideas are substantially different from initial ideas, and for 55.6\% we are able to increase novelty/creativity (meaning that, e.g., if 100 examples were updated, we would gain 56 examples that are more novel). The 2nd iteration,  further increases novelty for 57.8\% of the ideas that continued to another iteration. For ideas not considered more novel after applying our method, we do not observe a drop in novelty---the method either increases or maintains novelty.

Ideas after novelty iterations are longer than initial ideas. We examine the new terms added after filtering 359 words, including stopwords, as many generic words and terms are often added (e.g., ``novel model/method/approach'').  While our method helps boost novelty, overall the model often tends to suggest combinations between popular concepts (§\ref{sec:err}). Novelty boosting seemed to often focus on adding dynamic/adaptive modeling, graph models and representations, the fusion of multiple modalities and sources---and sometimes all at once (e.g., ``\textit{Dynamic Syntax-Aware Graph Fusion Networks (DSAGFN)}''), and to explicitly compare against existing ideas from literature (Table \ref{tab:GPT4}).

\begin{table}[!htb]
\centering
\small
\begin{tabularx}{\linewidth}{>{\hsize=2.2\hsize}X>{\centering\arraybackslash\hsize=1\hsize}X>{\centering\arraybackslash\hsize=0.6\hsize}X>{\centering\arraybackslash\hsize=0.6\hsize}X>{\centering\arraybackslash\hsize=0.6\hsize}X}
\toprule
\textbf{Type} &\texttt{Meditron}& \texttt{+SN}& \texttt{+CT}& \texttt{+KG}  \\ 
\midrule
Helpful(\%)    &  35  & 80  & 60 & 50 \\
Unhelpful(\%)  &  65  & 20  & 40 & 50 \\
vs. GT(\%)     &  30  &  45 & 50 & 35 \\
\bottomrule
\end{tabularx}
\caption{Human evaluations results of each system output for the idea sentence prediction task on Biomedical Domain. 
``\texttt{vs. GT}'' refers to percents which system outputs are better than ground truth ideas.
\label{tab:biohuman_evaluation}
}
\end{table}

\begin{table*}[!htb]
\centering
\small
\begin{tabularx}{\linewidth}{>{\hsize=0.2\hsize}X>{\arraybackslash\hsize=1.8\hsize}X}
\toprule
\textbf{Type}&     \textbf{Content}   \\
\midrule
Input   \cite{dong2022learning}                   & \textit{seed term}: \textbf{speech unit boundaries} ; \textit{context} (abridged): ... generate partial sentence translation given a streaming speech input. existing approaches ... break the acoustic units in speech, as boundaries between acoustic units in speech are not even.
...\\\hdashline
Initial idea     &  A pause prediction model to identify \textbf{speech unit boundaries} ...\\\hdashline  

Iteration 1    & A method that leverages acoustic and linguistic features to predict \textbf{speech unit boundaries} dynamically, ensuring smooth transitions ... differs from the existing research as it combines both acoustic properties and linguistic context ... adapting to variations in speaker characteristics, speaking styles, and languages.  \\\hdashline  

Iteration 2   &  A novel method called Adaptive \textbf{Speech Unit Boundary} Detection (ASUBD) ... a combination of attention mechanisms to focus on relevant acoustic and linguistic features and reinforcement learning to guide the system to make optimal predictions of unit boundaries based on previous decisions...
\\
\hdashline  

Ground Truth  &  ... an efficient monotonic segmentation module ... accumulate acoustic information incrementally and detect proper \textbf{speech unit boundaries}.\\

\bottomrule
\end{tabularx}
\caption{Example of iterative novelty iterations. Our novelty iteration method enhances ideas overall; however ideas are often based on superficial recombinations of common concepts, far from the technical depth of scientific papers.
\label{tab:GPT4}
}
\end{table*}

\subsubsection{Domain Generalization Case Study} 
Our domain-agnostic framework can be applied to other domains by changing the IE system used in the preprocessing procedure. To demonstrate this, we conduct an additional initial experiment in the biochemical domain. We follow a similar data creation procedure as for NLP papers. We collect a dataset from PubMed papers and use PubTator 3~\cite{Islamaj2021,wei2022tmvar,aioner,gnorm2,lai2023biorex} as an IE system to extract a KG from paper abstracts. We use a sentence classifier trained on annotated abstracts~\cite{huang-etal-2020-coda} to select background context. We fine-tune a state-of-the-art biomedical large language model~\cite{chen2023meditron} on our data and evaluate on a test split past its pre-training cutoff date.\footnote{More data and training details in Appendix~\ref{sec:bio_data}, \ref{app:biodetail}.} 
We ask two biochemical domain experts with graduate-level education to evaluate the quality of the results as before, finding them to overall rate 80\% of the generated directions positively. Finally, in contrast to NLP-domain experiments, evaluators were more satisfied with the generated outputs than the ground truth regarding technical detail. Detailed results are in Table~\ref{tab:biohuman_evaluation}. However, this preliminary experiment was meant mainly to demonstrate the generality of our approach, and a more in-depth exploration of utility and quality is left for future work.


\subsection{Error Analysis} 
\label{sec:err}
Models often made generic suggestions, woven together with specific details copied directly from the context (e.g., ``\textit{NLP with ML algorithms and sentiment analysis}'' for some problem X, or ``\textit{data augmentation and transfer learning}'' for Y, or ``\textit{BERT or RoBERTa}'' for Z). Our techniques reduced this behavior but did not fully solve it. GPT4 models, especially, seemed to generate generic descriptions of common steps in NLP workflows (e.g., ``\textit{Data preprocessing: Clean the text data, remove unnecessary characters, perform tokenization...}''). All models often copied and rephrased directly from the context. In certain cases, models applied simple logical modifications to the context; e.g., when contexts described problems such as ``\textit{high latency}'' or ``\textit{efficiency limitations}'', the suggestions would include phrases such as ``\textit{low latency}'' or ``\textit{highly efficient}''.

\subsection{Automated Evaluation Analysis} 
In open-ended tasks such as ours, automatic evaluations comparing system output to ground truth texts may be limited. Nonetheless, automated metrics such as ROUGE~\cite{lin-2004-rouge}, BERTScore~\cite{Zhang*2020BERTScore} and BARTScore \cite{yuan2021bartscore}, that check the similarity between ground truth and generated output, may surface interesting findings.
We find \texttt{GPT}-based models to be outperformed by \texttt{T5}-based models; \texttt{GPT4} outputs are much longer than \texttt{T5}, explaining why they underperform in automatic metrics but outperform in human evaluations (\S\ref{sec:human}). Generated sentences often follow certain templates (e.g., \textit{``In this paper, we propose a new ... for ...''}), which also helps explain why \texttt{T5} fine-tuned on many examples scores higher superficially. At the same time, our in-context contrastive examples which encourage novelty with respect to background context, helped models perform better than baseline fine-tuning by reducing reliance on copying. See results in Table~\ref{tab:sp_ncs} (Appendix \ref{sec:evalmetrics}).

\section{Conclusions and Future Directions}
We propose a new setting, model and comprehensive evaluation for scientific hypothesis generation with language models that are grounded in literature and optimized for novelty. We present a new framework named \textsc{SciMON} 
in which models take background problem contexts and provide suggestions that are novel while based on literature. Models retrieve inspirations from semantic similarity graphs, knowledge graphs, and citation networks. We introduce a new iterative novelty boosting mechanism that helps large language models (LLMs) such as GPT-4 generate more novel ideas by explicitly comparing ideas to prior work and refining them. Our experiments demonstrate that the task of generating natural language scientific hypotheses is highly challenging. While our methods improve upon baseline LLMs, generated ideas tend to be incremental and with insufficient detail. Generating novel and meaningful scientific concepts and their compositions remains a fundamental problem \cite{hope2022computational}. Evaluation in this setting is also highly challenging, with a huge space of potentially plausible hypotheses formulated in natural language. 
One interesting direction is to expand \textsc{SciMON} with a multimodal analysis of formulas, tables, and figures to provide a more comprehensive background context. 

\section{Limitations}
\label{sec:limitation}
We discuss limitations extensively throughout the paper, such as in terms of evaluation challenges and data quality. Here we include additional details on limitations. 

\subsection{Limitations of Data Collection}
We crawled papers with Semantic Scholar Academic Graph API from 1952 to June 2022. The number of available papers is limited by the data we crawled from the Semantic Scholar Academic Graph. We also crawled papers from PubMed 1988 to 2024/01. We remove papers that are not English. We also remove papers where abstracts are not correctly parsed from paper PDFs. We will expand our models to papers written in other languages and other domains in the future.

\subsection{Limitations of System Performance}
Our dataset is based on state-of-the-art IE systems, which may be noisy. For instance, the coreference and SciSpacy abbreviation resolution models fail to link \textit{A2LCTC} to \textit{Action-to-Language Connectionist Temporal Classification}. The background context detection may also have errors: e.g., the sentence classification component fails to treat \textit{``For example, the language models are overall more positive towards the stock market, but there are significant differences in preferences between a pair of industry sectors, or even within a sector.''} as background context. In our human-vetted gold data subset, we make sure to filter such cases, but they remain in the training data. SentenceBert \cite{reimers-gurevych-2019-sentence}, and GPT3.5/4 are not finetuned and might be biased towards pretraining datasets. 
The idea novelty boosting method is limited by the quality of retrieval models. Better retrieval models may be explored in the future. Due to hardware constraints, we mainly investigated models with up to 7 billion parameters. Due to API change and model randomness, our GPT3.5/4 results might not be easily reproducible.

\subsection{Limitations of Evaluation}
We recruit annotators from Ph.D. students; their opinions may differ from annotators who have different levels of domain knowledge. Our setting uses a seed term taken from the ground truth as input,  to emulate a scenario where a human provides guidance to an assistant model. Future work could explore methods in the setting without a seed term, an even harder task, or evaluate in an interactive setting with user-provided seed terms. In addition, while the seed is sampled from the ground truth, in our human-annotated gold subset, we make sure that in no case does the input context trivially leak the output.

\subsection{Memorization Check}
\citet{carlini2023quantifying} reports that LLMs tend to memorize part of their training data, a well-known concern in evaluating current LLMs. Therefore, we examine the pretraining data of each model:

\begin{itemize}
 \item T5: \citet{JMLR:v21:20-074} shows that T5 is pretrained on C4 which was crawled from web prior to April 2019.
 \item GPT3.5: Based on the documentation,\footnote{\url{platform.openai.com/docs/model-index-for-researchers}} GPT-3.5 series is pretrained on a combination of test and code from before Q4 2021. 
 \item GPT4: \citet{openai2023gpt4} shows that the GPT-4 checkpoint we used utilizes most pertaining data before September 2021. Despite this, the pretraining and post-training data contain ``a small amount'' of more recent data.\footnote{See footnote 10, page 10 of \citet{openai2023gpt4}.}

\end{itemize}

Because we evaluate our models on papers published in 2022, the likelihood of test papers appearing in the pretraining corpora for the models is substantially reduced. We additionally performed a manual examination of GPT-4 memorization in our gold set based on 2022 ACL Anthology papers, by seeing if GPT-4 could complete information such as method names or generate text that strongly mimics the ground truth papers, and found no evidence of this occurring. 
The Meditron-7b~\cite{chen2023meditron} uses PubMed with a cut-off in August 2023, and our biochemical test set only includes PubMed papers after 2023/08.

\section*{Acknowledgements}
 This work is supported by the Molecule Maker Lab Institute: an AI research institute program supported by NSF under award No. 2019897, by DOE Center for Advanced Bioenergy and Bioproducts Innovation U.S. Department of Energy, Office of Science, Office of Biological and Environmental Research under Award Number DESC0018420, by U.S. the AI Research Institutes program by National Science Foundation and the Institute of Education Sciences, Department of Education through Award No. 2229873 - AI Institute for Transforming Education for Children with Speech and Language Processing Challenges, and by AI Agriculture: the Agriculture and Food Research Initiative (AFRI) grant no. 2020-67021- 32799/project accession no.1024178 from the USDA National Institute of Food and Agriculture. The views and conclusions contained herein are those of the authors and should not be interpreted as necessarily representing the official policies, either expressed or implied of, the National Science Foundation, the U.S. Department of Energy, and the U.S. Government. The U.S. Government is authorized to reproduce and distribute reprints for governmental purposes notwithstanding any copyright annotation therein.
\bibliography{anthology,custom}

\begin{thebibliography}{48}
\expandafter\ifx\csname natexlab\endcsname\relax\def\natexlab#1{#1}\fi

\bibitem[{Boiko et~al.(2023)Boiko, MacKnight, Kline, and
  Gomes}]{boiko2023autonomous}
Daniil~A Boiko, Robert MacKnight, Ben Kline, and Gabe Gomes. 2023.
\newblock Autonomous chemical research with large language models.
\newblock \emph{Nature}, 624(7992):570--578.

\bibitem[{Brown et~al.(2020)Brown, Mann, Ryder, Subbiah, Kaplan, Dhariwal,
  Neelakantan, Shyam, Sastry, Askell, Agarwal, Herbert-Voss, Krueger, Henighan,
  Child, Ramesh, Ziegler, Wu, Winter, Hesse, Chen, Sigler, Litwin, Gray, Chess,
  Clark, Berner, McCandlish, Radford, Sutskever, and
  Amodei}]{NEURIPS2020_1457c0d6}
Tom Brown, Benjamin Mann, Nick Ryder, Melanie Subbiah, Jared~D Kaplan, Prafulla
  Dhariwal, Arvind Neelakantan, Pranav Shyam, Girish Sastry, Amanda Askell,
  Sandhini Agarwal, Ariel Herbert-Voss, Gretchen Krueger, Tom Henighan, Rewon
  Child, Aditya Ramesh, Daniel Ziegler, Jeffrey Wu, Clemens Winter, Chris
  Hesse, Mark Chen, Eric Sigler, Mateusz Litwin, Scott Gray, Benjamin Chess,
  Jack Clark, Christopher Berner, Sam McCandlish, Alec Radford, Ilya Sutskever,
  and Dario Amodei. 2020.
\newblock \href
  {https://proceedings.neurips.cc/paper/2020/file/1457c0d6bfcb4967418bfb8ac142f64a-Paper.pdf}
  {Language models are few-shot learners}.
\newblock In \emph{Advances in Neural Information Processing Systems},
  volume~33, pages 1877--1901. Curran Associates, Inc.

\bibitem[{Carlini(2023)}]{carlini2023llm}
Nicholas Carlini. 2023.
\newblock \href {http://arxiv.org/abs/2307.15008} {A llm assisted exploitation
  of ai-guardian}.
\newblock \emph{Cryptography and Security Repository}, arXiv:2307.15008.

\bibitem[{Carlini et~al.(2023)Carlini, Ippolito, Jagielski, Lee, Tramer, and
  Zhang}]{carlini2023quantifying}
Nicholas Carlini, Daphne Ippolito, Matthew Jagielski, Katherine Lee, Florian
  Tramer, and Chiyuan Zhang. 2023.
\newblock \href {https://openreview.net/forum?id=TatRHT_1cK} {Quantifying
  memorization across neural language models}.
\newblock In \emph{The Eleventh International Conference on Learning
  Representations}.

\bibitem[{Cattan et~al.(2021)Cattan, Johnson, Weld, Dagan, Beltagy, Downey, and
  Hope}]{cattan2021scico}
Arie Cattan, Sophie Johnson, Daniel~S Weld, Ido Dagan, Iz~Beltagy, Doug Downey,
  and Tom Hope. 2021.
\newblock \href {https://openreview.net/forum?id=OFLbgUP04nC} {Scico:
  Hierarchical cross-document coreference for scientific concepts}.
\newblock In \emph{3rd Conference on Automated Knowledge Base Construction}.

\bibitem[{Chen et~al.(2023)Chen, Cano, Romanou, Bonnet, Matoba, Salvi,
  Pagliardini, Fan, K{\"o}pf, Mohtashami et~al.}]{chen2023meditron}
Zeming Chen, Alejandro~Hern{\'a}ndez Cano, Angelika Romanou, Antoine Bonnet,
  Kyle Matoba, Francesco Salvi, Matteo Pagliardini, Simin Fan, Andreas
  K{\"o}pf, Amirkeivan Mohtashami, et~al. 2023.
\newblock \href {http://arxiv.org/abs/2311.03748} {Meditron-70b: Scaling
  medical pretraining for large language models}.
\newblock \emph{Computation and Language Repository}, arXiv:2311.16079.

\bibitem[{Cohan et~al.(2019)Cohan, Beltagy, King, Dalvi, and
  Weld}]{cohan-etal-2019-pretrained}
Arman Cohan, Iz~Beltagy, Daniel King, Bhavana Dalvi, and Dan Weld. 2019.
\newblock \href {https://doi.org/10.18653/v1/D19-1383} {Pretrained language
  models for sequential sentence classification}.
\newblock In \emph{Proceedings of the 2019 Conference on Empirical Methods in
  Natural Language Processing and the 9th International Joint Conference on
  Natural Language Processing (EMNLP-IJCNLP)}, pages 3693--3699, Hong Kong,
  China. Association for Computational Linguistics.

\bibitem[{de~Masson~d\textquotesingle Autume
  et~al.(2019)de~Masson~d\textquotesingle Autume, Ruder, Kong, and
  Yogatama}]{NEURIPS2019_f8d2e80c}
Cyprien de~Masson~d\textquotesingle Autume, Sebastian Ruder, Lingpeng Kong, and
  Dani Yogatama. 2019.
\newblock \href
  {https://proceedings.neurips.cc/paper/2019/file/f8d2e80c1458ea2501f98a2cafadb397-Paper.pdf}
  {Episodic memory in lifelong language learning}.
\newblock In \emph{Advances in Neural Information Processing Systems},
  volume~32. Curran Associates, Inc.

\bibitem[{Dong et~al.(2022)Dong, Zhu, Wang, and Li}]{dong2022learning}
Qian Dong, Yaoming Zhu, Mingxuan Wang, and Lei Li. 2022.
\newblock Learning when to translate for streaming speech.
\newblock In \emph{Proceedings of the 60th Annual Meeting of the Association
  for Computational Linguistics (Volume 1: Long Papers)}, pages 680--694.

\bibitem[{Henry and McInnes(2017)}]{henry2017literature}
Sam Henry and Bridget~T McInnes. 2017.
\newblock Literature based discovery: models, methods, and trends.
\newblock \emph{Journal of biomedical informatics}, 74:20--32.

\bibitem[{Hope et~al.(2023)Hope, Downey, Etzioni, Weld, and
  Horvitz}]{hope2022computational}
Tom Hope, Doug Downey, Oren Etzioni, Daniel~S Weld, and Eric Horvitz. 2023.
\newblock \href {https://arxiv.org/abs/2205.02007} {A computational inflection
  for scientific discovery}.
\newblock \emph{Communications of the ACM}.

\bibitem[{Hu et~al.(2021)Hu, Fu, Yin, and de~Melo}]{hu-etal-2021-context}
Zhe Hu, Zuohui Fu, Yu~Yin, and Gerard de~Melo. 2021.
\newblock \href {https://doi.org/10.18653/v1/2021.emnlp-main.312}
  {Context-aware interaction network for question matching}.
\newblock In \emph{Proceedings of the 2021 Conference on Empirical Methods in
  Natural Language Processing}, pages 3846--3853, Online and Punta Cana,
  Dominican Republic. Association for Computational Linguistics.

\bibitem[{Huang et~al.(2023)Huang, Vora, Liang, and
  Leskovec}]{huang2023benchmarking}
Qian Huang, Jian Vora, Percy Liang, and Jure Leskovec. 2023.
\newblock \href {http://arxiv.org/abs/2310.03302} {Benchmarking large language
  models as ai research agents}.
\newblock \emph{Machine Learning Repository}, arXiv:2310.03302.

\bibitem[{Huang et~al.(2020)Huang, Huang, Ding, Hsu, and
  Giles}]{huang-etal-2020-coda}
Ting-Hao~Kenneth Huang, Chieh-Yang Huang, Chien-Kuang~Cornelia Ding, Yen-Chia
  Hsu, and C.~Lee Giles. 2020.
\newblock \href {https://aclanthology.org/2020.nlpcovid19-acl.6} {{CODA-19}:
  Using a non-expert crowd to annotate research aspects on 10,000+ abstracts in
  the {COVID-19} open research dataset}.
\newblock In \emph{Proceedings of the 1st Workshop on {NLP} for {COVID-19} at
  {ACL} 2020}, Online. Association for Computational Linguistics.

\bibitem[{Islamaj et~al.(2021)Islamaj, Leaman, Kim, Kwon, Wei, Comeau, Peng,
  Cissel, Coss, Fisher, Guzman, Kochar, Koppel, Trinh, Sekiya, Ward, Whitman,
  Schmidt, and Lu}]{Islamaj2021}
Rezarta Islamaj, Robert Leaman, Sun Kim, Dongseop Kwon, Chih-Hsuan Wei,
  Donald~C. Comeau, Yifan Peng, David Cissel, Cathleen Coss, Carol Fisher, Rob
  Guzman, Preeti~Gokal Kochar, Stella Koppel, Dorothy Trinh, Keiko Sekiya,
  Janice Ward, Deborah Whitman, Susan Schmidt, and Zhiyong Lu. 2021.
\newblock \href {https://doi.org/10.1038/s41597-021-00875-1} {Nlm-chem, a new
  resource for chemical entity recognition in pubmed full text literature}.
\newblock \emph{Scientific Data}, 8(1):91.

\bibitem[{Jimenez~Gutierrez et~al.(2022)Jimenez~Gutierrez, McNeal, Washington,
  Chen, Li, Sun, and Su}]{jimenez-gutierrez-etal-2022-thinking}
Bernal Jimenez~Gutierrez, Nikolas McNeal, Clayton Washington, You Chen, Lang
  Li, Huan Sun, and Yu~Su. 2022.
\newblock \href {https://aclanthology.org/2022.findings-emnlp.329} {Thinking
  about {GPT}-3 in-context learning for biomedical {IE}? think again}.
\newblock In \emph{Findings of the Association for Computational Linguistics:
  EMNLP 2022}, pages 4497--4512, Abu Dhabi, United Arab Emirates. Association
  for Computational Linguistics.

\bibitem[{Lai et~al.(2023)Lai, Wei, Luo, Chen, and Lu}]{lai2023biorex}
Po-Ting Lai, Chih-Hsuan Wei, Ling Luo, Qingyu Chen, and Zhiyong Lu. 2023.
\newblock \href {https://doi.org/https://doi.org/10.1016/j.jbi.2023.104487}
  {Biorex: Improving biomedical relation extraction by leveraging heterogeneous
  datasets}.
\newblock \emph{Journal of Biomedical Informatics}, 146:104487.

\bibitem[{Lin(2004)}]{lin-2004-rouge}
Chin-Yew Lin. 2004.
\newblock \href {https://aclanthology.org/W04-1013} {{ROUGE}: A package for
  automatic evaluation of summaries}.
\newblock In \emph{Text Summarization Branches Out}, pages 74--81, Barcelona,
  Spain. Association for Computational Linguistics.

\bibitem[{Liu et~al.(2022)Liu, Shen, Zhang, Dolan, Carin, and
  Chen}]{liu-etal-2022-makes}
Jiachang Liu, Dinghan Shen, Yizhe Zhang, Bill Dolan, Lawrence Carin, and Weizhu
  Chen. 2022.
\newblock \href {https://doi.org/10.18653/v1/2022.deelio-1.10} {What makes good
  in-context examples for {GPT}-3?}
\newblock In \emph{Proceedings of Deep Learning Inside Out (DeeLIO 2022): The
  3rd Workshop on Knowledge Extraction and Integration for Deep Learning
  Architectures}, pages 100--114, Dublin, Ireland and Online. Association for
  Computational Linguistics.

\bibitem[{Lo et~al.(2020)Lo, Wang, Neumann, Kinney, and
  Weld}]{lo-etal-2020-s2orc}
Kyle Lo, Lucy~Lu Wang, Mark Neumann, Rodney Kinney, and Daniel Weld. 2020.
\newblock \href {https://doi.org/10.18653/v1/2020.acl-main.447} {{S}2{ORC}: The
  semantic scholar open research corpus}.
\newblock In \emph{Proceedings of the 58th Annual Meeting of the Association
  for Computational Linguistics}, pages 4969--4983, Online. Association for
  Computational Linguistics.

\bibitem[{Loshchilov and Hutter(2019)}]{Loshchilov2019DecoupledWD}
Ilya Loshchilov and Frank Hutter. 2019.
\newblock \href {https://openreview.net/pdf?id=Bkg6RiCqY7} {Decoupled weight
  decay regularization}.
\newblock In \emph{Proceedings of the 7th International Conference on Learning
  Representations}.

\bibitem[{Luan et~al.(2018)Luan, He, Ostendorf, and
  Hajishirzi}]{luan-etal-2018-multi}
Yi~Luan, Luheng He, Mari Ostendorf, and Hannaneh Hajishirzi. 2018.
\newblock \href {https://doi.org/10.18653/v1/D18-1360} {Multi-task
  identification of entities, relations, and coreference for scientific
  knowledge graph construction}.
\newblock In \emph{Proceedings of the 2018 Conference on Empirical Methods in
  Natural Language Processing}, pages 3219--3232, Brussels, Belgium.
  Association for Computational Linguistics.

\bibitem[{Luo et~al.(2023)Luo, Wei, Lai, Leaman, Chen, and Lu}]{aioner}
Ling Luo, Chih-Hsuan Wei, Po-Ting Lai, Robert Leaman, Qingyu Chen, and Zhiyong
  Lu. 2023.
\newblock \href {https://doi.org/10.1093/bioinformatics/btad310} {{AIONER:
  all-in-one scheme-based biomedical named entity recognition using deep
  learning}}.
\newblock \emph{Bioinformatics}, 39(5):btad310.

\bibitem[{Nadkarni et~al.(2021)Nadkarni, Wadden, Beltagy, Smith, Hajishirzi,
  and Hope}]{nadkarni2021scientific}
Rahul Nadkarni, David Wadden, Iz~Beltagy, Noah~A Smith, Hannaneh Hajishirzi,
  and Tom Hope. 2021.
\newblock Scientific language models for biomedical knowledge base completion:
  an empirical study.
\newblock \emph{AKBC}.

\bibitem[{Neumann et~al.(2019)Neumann, King, Beltagy, and
  Ammar}]{neumann-etal-2019-scispacy}
Mark Neumann, Daniel King, Iz~Beltagy, and Waleed Ammar. 2019.
\newblock \href {https://doi.org/10.18653/v1/W19-5034} {{S}cispa{C}y: Fast and
  robust models for biomedical natural language processing}.
\newblock In \emph{Proceedings of the 18th BioNLP Workshop and Shared Task},
  pages 319--327, Florence, Italy. Association for Computational Linguistics.

\bibitem[{Newell and Simon(1956)}]{newell1956logic}
Allen Newell and Herbert Simon. 1956.
\newblock The logic theory machine--a complex information processing system.
\newblock \emph{IRE Transactions on information theory}, 2(3):61--79.

\bibitem[{Nori et~al.(2023)Nori, King, McKinney, Carignan, and
  Horvitz}]{nori2023capabilities}
Harsha Nori, Nicholas King, Scott~Mayer McKinney, Dean Carignan, and Eric
  Horvitz. 2023.
\newblock \href {http://arxiv.org/abs/2303.13375} {Capabilities of {GPT}-4 on
  medical challenge problems}.
\newblock \emph{Computation and Language Repository}, arXiv:2303.13375.

\bibitem[{Oord et~al.(2018)Oord, Li, and Vinyals}]{oord2018representation}
Aaron van~den Oord, Yazhe Li, and Oriol Vinyals. 2018.
\newblock \href {http://arxiv.org/abs/1807.03748} {Representation learning with
  contrastive predictive coding}.
\newblock \emph{Machine Learning Repository}, arXiv:1807.03748.

\bibitem[{OpenAI(2023)}]{openai2023gpt4}
OpenAI. 2023.
\newblock \href {https://arxiv.org/abs/2303.08774} {Gpt-4 technical report}.
\newblock \emph{Computation and Language Repository}, arXiv:2303.08774.

\bibitem[{Ouyang et~al.(2022)Ouyang, Wu, Jiang, Almeida, Wainwright, Mishkin,
  Zhang, Agarwal, Slama, Ray et~al.}]{ouyang2022training}
Long Ouyang, Jeffrey Wu, Xu~Jiang, Diogo Almeida, Carroll Wainwright, Pamela
  Mishkin, Chong Zhang, Sandhini Agarwal, Katarina Slama, Alex Ray, et~al.
  2022.
\newblock \href
  {https://cdn.openai.com/papers/Training_language_models_to_follow_instructions_with_human_feedback.pdf}
  {Training language models to follow instructions with human feedback}.
\newblock \emph{Advances in Neural Information Processing Systems},
  35:27730--27744.

\bibitem[{Qin et~al.(2022)Qin, Zhang, Lin, Liu, Li, Sun, and
  Zhou}]{qin-etal-2022-elle}
Yujia Qin, Jiajie Zhang, Yankai Lin, Zhiyuan Liu, Peng Li, Maosong Sun, and Jie
  Zhou. 2022.
\newblock \href {https://doi.org/10.18653/v1/2022.findings-acl.220} {{ELLE}:
  Efficient lifelong pre-training for emerging data}.
\newblock In \emph{Findings of the Association for Computational Linguistics:
  ACL 2022}, pages 2789--2810, Dublin, Ireland. Association for Computational
  Linguistics.

\bibitem[{Raffel et~al.(2020)Raffel, Shazeer, Roberts, Lee, Narang, Matena,
  Zhou, Li, and Liu}]{JMLR:v21:20-074}
Colin Raffel, Noam Shazeer, Adam Roberts, Katherine Lee, Sharan Narang, Michael
  Matena, Yanqi Zhou, Wei Li, and Peter~J. Liu. 2020.
\newblock \href {http://jmlr.org/papers/v21/20-074.html} {Exploring the limits
  of transfer learning with a unified text-to-text transformer}.
\newblock \emph{Journal of Machine Learning Research}, 21(140):1--67.

\bibitem[{Reimers and Gurevych(2019)}]{reimers-gurevych-2019-sentence}
Nils Reimers and Iryna Gurevych. 2019.
\newblock \href {https://doi.org/10.18653/v1/D19-1410} {Sentence-{BERT}:
  Sentence embeddings using {S}iamese {BERT}-networks}.
\newblock In \emph{Proceedings of the 2019 Conference on Empirical Methods in
  Natural Language Processing and the 9th International Joint Conference on
  Natural Language Processing (EMNLP-IJCNLP)}, pages 3982--3992, Hong Kong,
  China. Association for Computational Linguistics.

\bibitem[{Simon(1973)}]{simon1973does}
Herbert~A Simon. 1973.
\newblock Does scientific discovery have a logic?
\newblock \emph{Philosophy of science}, 40(4):471--480.

\bibitem[{Sosa and Altman(2022)}]{sosa2022contexts}
Daniel~N Sosa and Russ~B Altman. 2022.
\newblock \href {https://pubmed.ncbi.nlm.nih.gov/35817308/} {Contexts and
  contradictions: a roadmap for computational drug repurposing with knowledge
  inference}.
\newblock \emph{Briefings in Bioinformatics}, 23(4):bbac268.

\bibitem[{Swanson(1986)}]{swanson1986undiscovered}
Don~R Swanson. 1986.
\newblock \href {https://www.jstor.org/stable/4307965} {Undiscovered public
  knowledge}.
\newblock \emph{The Library Quarterly}, 56(2):103--118.

\bibitem[{Sybrandt et~al.(2020)Sybrandt, Tyagin, Shtutman, and
  Safro}]{sybrandt2020agatha}
Justin Sybrandt, Ilya Tyagin, Michael Shtutman, and Ilya Safro. 2020.
\newblock \href {https://dl.acm.org/doi/10.1145/3340531.3412684} {Agatha:
  automatic graph mining and transformer based hypothesis generation approach}.
\newblock In \emph{Proceedings of the 29th ACM International Conference on
  Information \& Knowledge Management}, pages 2757--2764.

\bibitem[{Tshitoyan et~al.(2019)Tshitoyan, Dagdelen, Weston, Dunn, Rong,
  Kononova, Persson, Ceder, and Jain}]{tshitoyan2019unsupervised}
Vahe Tshitoyan, John Dagdelen, Leigh Weston, Alexander Dunn, Ziqin Rong, Olga
  Kononova, Kristin~A Persson, Gerbrand Ceder, and Anubhav Jain. 2019.
\newblock \href {https://www.nature.com/articles/s41586-019-1335-8}
  {Unsupervised word embeddings capture latent knowledge from materials science
  literature}.
\newblock \emph{Nature}, 571(7763):95--98.

\bibitem[{Wang et~al.(2019)Wang, Huang, Jiang, Knight, Ji, Bansal, and
  Luan}]{wang-etal-2019-paperrobot}
Qingyun Wang, Lifu Huang, Zhiying Jiang, Kevin Knight, Heng Ji, Mohit Bansal,
  and Yi~Luan. 2019.
\newblock \href {https://doi.org/10.18653/v1/P19-1191} {{P}aper{R}obot:
  Incremental draft generation of scientific ideas}.
\newblock In \emph{Proceedings of the 57th Annual Meeting of the Association
  for Computational Linguistics}, pages 1980--1991, Florence, Italy.
  Association for Computational Linguistics.

\bibitem[{Warstadt et~al.(2023)Warstadt, Mueller, Choshen, Wilcox, Zhuang,
  Ciro, Mosquera, Paranjabe, Williams, Linzen et~al.}]{warstadt2023findings}
Alex Warstadt, Aaron Mueller, Leshem Choshen, Ethan Wilcox, Chengxu Zhuang,
  Juan Ciro, Rafael Mosquera, Bhargavi Paranjabe, Adina Williams, Tal Linzen,
  et~al. 2023.
\newblock Findings of the babylm challenge: Sample-efficient pretraining on
  developmentally plausible corpora.
\newblock In \emph{Proceedings of the BabyLM Challenge at the 27th Conference
  on Computational Natural Language Learning}, pages 1--34.

\bibitem[{Wei et~al.(2022)Wei, Allot, Riehle, Milosavljevic, and
  Lu}]{wei2022tmvar}
Chih-Hsuan Wei, Alexis Allot, Kevin Riehle, Aleksandar Milosavljevic, and
  Zhiyong Lu. 2022.
\newblock \href {https://www.ncbi.nlm.nih.gov/pmc/articles/PMC9477515/} {tmvar
  3.0: an improved variant concept recognition and normalization tool}.
\newblock \emph{Bioinformatics}, 38(18):4449--4451.

\bibitem[{Wei et~al.(2023)Wei, Luo, Islamaj, Lai, and Lu}]{gnorm2}
Chih-Hsuan Wei, Ling Luo, Rezarta Islamaj, Po-Ting Lai, and Zhiyong Lu. 2023.
\newblock \href {https://doi.org/10.1093/bioinformatics/btad599} {{GNorm2: an
  improved gene name recognition and normalization system}}.
\newblock \emph{Bioinformatics}, 39(10):btad599.

\bibitem[{Wolf et~al.(2020)Wolf, Debut, Sanh, Chaumond, Delangue, Moi, Cistac,
  Rault, Louf, Funtowicz, Davison, Shleifer, von Platen, Ma, Jernite, Plu, Xu,
  Le~Scao, Gugger, Drame, Lhoest, and Rush}]{wolf-etal-2020-transformers}
Thomas Wolf, Lysandre Debut, Victor Sanh, Julien Chaumond, Clement Delangue,
  Anthony Moi, Pierric Cistac, Tim Rault, Remi Louf, Morgan Funtowicz, Joe
  Davison, Sam Shleifer, Patrick von Platen, Clara Ma, Yacine Jernite, Julien
  Plu, Canwen Xu, Teven Le~Scao, Sylvain Gugger, Mariama Drame, Quentin Lhoest,
  and Alexander Rush. 2020.
\newblock \href {https://doi.org/10.18653/v1/2020.emnlp-demos.6} {Transformers:
  State-of-the-art natural language processing}.
\newblock In \emph{Proceedings of the 2020 Conference on Empirical Methods in
  Natural Language Processing: System Demonstrations}, pages 38--45, Online.
  Association for Computational Linguistics.

\bibitem[{Xu et~al.(2023)Xu, Sheng, Xue, Fu, Wang, and
  Zhou}]{xu-etal-2023-exploring}
Yi~Xu, Shuqian Sheng, Bo~Xue, Luoyi Fu, Xinbing Wang, and Chenghu Zhou. 2023.
\newblock \href {https://doi.org/10.18653/v1/2023.acl-long.727} {Exploring and
  verbalizing academic ideas by concept co-occurrence}.
\newblock In \emph{Proceedings of the 61st Annual Meeting of the Association
  for Computational Linguistics (Volume 1: Long Papers)}, pages 13001--13027,
  Toronto, Canada. Association for Computational Linguistics.

\bibitem[{Ye et~al.(2022)Ye, Lin, Li, and Sun}]{ye-etal-2022-packed}
Deming Ye, Yankai Lin, Peng Li, and Maosong Sun. 2022.
\newblock \href {https://doi.org/10.18653/v1/2022.acl-long.337} {Packed
  levitated marker for entity and relation extraction}.
\newblock In \emph{Proceedings of the 60th Annual Meeting of the Association
  for Computational Linguistics (Volume 1: Long Papers)}, pages 4904--4917,
  Dublin, Ireland. Association for Computational Linguistics.

\bibitem[{Yuan et~al.(2021)Yuan, Neubig, and Liu}]{yuan2021bartscore}
Weizhe Yuan, Graham Neubig, and Pengfei Liu. 2021.
\newblock \href {https://openreview.net/forum?id=5Ya8PbvpZ9} {{BARTS}core:
  Evaluating generated text as text generation}.
\newblock In \emph{Advances in Neural Information Processing Systems}.

\bibitem[{Zhang* et~al.(2020)Zhang*, Kishore*, Wu*, Weinberger, and
  Artzi}]{Zhang*2020BERTScore}
Tianyi Zhang*, Varsha Kishore*, Felix Wu*, Kilian~Q. Weinberger, and Yoav
  Artzi. 2020.
\newblock \href {https://openreview.net/forum?id=SkeHuCVFDr} {Bertscore:
  Evaluating text generation with bert}.
\newblock In \emph{Proceedings of the 8th International Conference on Learning
  Representations}.

\bibitem[{Zhou et~al.(2022)Zhou, Li, He, Bing, Cambria, Si, and
  Miao}]{zhou-etal-2022-melm}
Ran Zhou, Xin Li, Ruidan He, Lidong Bing, Erik Cambria, Luo Si, and Chunyan
  Miao. 2022.
\newblock \href {https://doi.org/10.18653/v1/2022.acl-long.160} {{MELM}: Data
  augmentation with masked entity language modeling for low-resource {NER}}.
\newblock In \emph{Proceedings of the 60th Annual Meeting of the Association
  for Computational Linguistics (Volume 1: Long Papers)}, pages 2251--2262,
  Dublin, Ireland. Association for Computational Linguistics.

\end{thebibliography}
\bibliographystyle{acl_natbib}
\appendix
\section{Dataset Collection}
\subsection{NLP Dataset Collection}
\label{sec:data}
\begin{figure*}[htb]
\centering
\includegraphics[width=\linewidth]{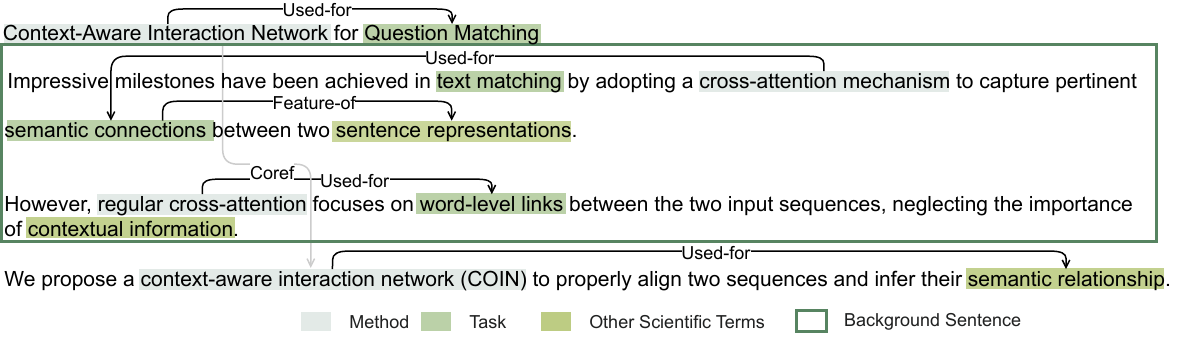}
\caption{\label{fig:IE} Preprocessing result for  \citet{hu-etal-2021-context} in non-canonicalized KG Corpus }
\end{figure*}

We download ACL Anthology papers from 1952 to 2022 using Semantic Scholar Academic Graph API.\footnote{\url{www.semanticscholar.org/product/api}} We filter out papers without abstracts and not written in English to obtain 67,408 papers.
Our dataset has 58,874 papers before 2021, 5,946 papers from 2021, and 2,588 from 2022. We first use PL-Marker~\cite{ye-etal-2022-packed} pretrained on SciERC~\cite{luan-etal-2018-multi} to extract nodes belonging to six types: \textit{Task}, \textit{Method}, \textit{Evaluation Metric}, \textit{Material}, \textit{Other Scientific Terms}, and \textit{Generic Terms}. The model then predicts relations between nodes belonging to seven relation types: \textit{Used-for}, \textit{Feature-of}, \textit{Evaluate-for}, \textit{Hyponym-of}, \textit{Part-of}, \textit{Compare}, and \textit{Conjunction}. Because we want to generate new ideas, we focus on \textit{used-for} relations in papers. Next, we use SciCo~\cite{cattan2021scico} with checkpoint from Hugging Face\footnote{\url{huggingface.co/allenai/longformer-scico}} to obtain entity coreference to merge identical nodes. Then, we use ScispaCy~\cite{neumann-etal-2019-scispacy} to perform unsupervised abbreviation detection to replace the abbreviation with a more informative long form. Finally, we perform scientific sentence classification~\cite{cohan-etal-2019-pretrained}\footnote{\url{github.com/allenai/sequential_sentence_classification}} to classify sentences from the abstract into five categories including \textit{Background}, \textit{Method}, \textit{Objective}, \textit{Other}, and \textit{Result}. We select sentences with labels of \textit{Background} and \textit{Other} as background context. 
During preprocessing, we only keep high-confidence outputs from IE models.
Figure~\ref{fig:IE} shows an example of the IE systems pipeline.

 \begin{table*}[!htb]
\centering
\small
\begin{tabularx}{\linewidth}{>{\hsize=0.3\hsize}X>{\centering\arraybackslash\hsize=1\hsize}X>{\centering\arraybackslash\hsize=1.1\hsize}X>{\centering\arraybackslash\hsize=1.1\hsize}X>{\centering\arraybackslash\hsize=1.4\hsize}X>{\centering\arraybackslash\hsize=1.1\hsize}X}
\toprule
\textbf{Stage}    & \textbf{ PL-Maker Entities}  &\textbf{PL-Maker Used-for Relations} & \textbf{SciCo Coreference}   & \textbf{Scispacy Abbreviation Detection}& \textbf{Sentence Classification}\\
\midrule
Precision &91.3	&65.4	&97.2	&100	&100\\
\bottomrule
\end{tabularx}
\caption{Human quality evaluation of preprocessing stages(\%). Overall pass rate after all steps are applied is 79.7\%. \label{tab:hum_pre}}
\end{table*}

\subsection{Biochemical Dataset Collection}

\label{sec:bio_data}

We collect PubMed papers from 1988 to 2024 using Entrez Programming Utilities API\footnote{\url{www.ncbi.nlm.nih.gov/books/NBK25501/}} for the following topics, including \textit{Yarrowia}, \textit{Saccharomyces cerevisiae}, \textit{Issatchenkia orientalis}, and \textit{Rhodosporidium toruloides}. We use PubTator 3~\cite{Islamaj2021,wei2022tmvar,aioner,gnorm2,lai2023biorex}. The PubTator 3 performs named entity recognition, relation extraction, entity coreference and linking, and entity normalization for the abstracts in the dataset. PubTator 3 identifies bio entities belonging to seven types:  \textit{gene}, \textit{chemical}, \textit{chromosome}, \textit{cell line}, \textit{variant}, \textit{disease}, and \textit{species}l and relations belonging to 13 types: \textit{associate}, \textit{cause}, \textit{compare}, \textit{convert}, \textit{contract}, \textit{drug interact}, \textit{inhibit}, \textit{interact}, \textit{negative correlate}, \textit{positive correlate}, \textit{prevent}, \textit{stimulate}, and \textit{treat}. Finally, we use a sentence classifier trained on CODA-19~\cite{huang-etal-2020-coda} to classify sentences in abstracts into \textit{background}, \textit{purpose}, \textit{method}, \textit{finding}, and \textit{other}. We select sentences with labels of \textit{background} as background context and remove sentences with labels of \textit{other}. We treat the rest sentences that have at least one entity as the target sentence. We only keep samples with low similarity between background context and corresponding ground truth sentences.\footnote{The similarity is calculated with \texttt{all-mpnet-base-v2}.} 
Our final dataset has 4,767 papers before 2023/02, 642 papers from 2023/02 to 2023/08, and 299 papers after 2023/08.

\section{Finetuning and Automated Evaluation details}
\label{sec:ft}
\subsection{Inspiration Retrieval Module}
The statistics of each inspiration type are in Table~\ref{tab:kg_nb}. Table~\ref{tab:neigh_s} shows sample retrieved inspirations. 

\begin{table}[!htb]
\centering
\small
\begin{tabularx}{\linewidth}{>{\hsize=1\hsize}X>{\centering\arraybackslash\hsize=1\hsize}X>{\centering\arraybackslash\hsize=1\hsize}X>{\centering\arraybackslash\hsize=1\hsize}X}
\toprule
\textbf{Type}&\textbf{Train}    &   \textbf{Valid}  &\textbf{Test} \\
\midrule
SN&  10.8  & 10.0  & 10.0\\
KG&  8.3  &  8.0 & 8.1\\
CT&  4.9  &  5.0 & 5.0\\
\bottomrule
\end{tabularx}
\caption{Average of \# of neighbors for each instance, excluding those which do not have any neighbor\label{tab:kg_nb}}
\end{table}

\subsubsection{Semantic Neighbors}  
We use \texttt{all-mpnet-base-v2} from SentenceBert \citep{reimers-gurevych-2019-sentence}, which performs best in semantic search to retrieve similar nodes from the training set based on query $q$ in \S\ref{sec:retrieval}. We retrieve up to 20 relevant semantic neighbors $\mathcal{R}$ from the training set for each instance. We treat the target nodes from $\mathcal{R}$ as semantic neighbors. 

\subsubsection{KG Neighbors} 
We use one-hop connected neighbors from the background KG $\mathcal{G}_B$ constructed on papers before 2021(i.e., the papers in the training set). Because of the scarcity of KG neighbors, we do not limit the number of KG neighbors.

\subsubsection{Citation Neighbors} 
Similar to semantic neighbors, we use $\mathtt{all-mpnet-base-v2}$ from SentenceBert \citep{reimers-gurevych-2019-sentence} to retrieve cited paper titles similar to query $q$. We restrict cited papers only before 2021. We retrieve up to 5 relevant citation neighbors from the papers' citation network. 

\subsection{Generation Module}
Our T5 model
and their variants are built based on the Huggingface framework \citep{wolf-etal-2020-transformers}.\footnote{\url{github.com/huggingface/transformers}}  We optimize those models by AdamW \citep{Loshchilov2019DecoupledWD} with the linear warmup scheduler.\footnote{\url{huggingface.co/docs/transformers/main\_classes/optimizer\_schedules\#transformers.get\_linear\_schedule\_with\_warmup}} Those models are finetuned on 4 NVIDIA A6000 48GB GPUs with distributed data parallel.\footnote{\url{pytorch.org/tutorials/intermediate/ddp\_tutorial.html}} The training time for each model is about 10 hours.

\subsubsection{In-Context Learning}
We choose GPT3.5 \texttt{davinci-003}\footnote{\url{openai.com/api/}} \cite{NEURIPS2020_1457c0d6} as our out-of-the-box causal language modeling baseline. We select $5$ instances from the training set as examples for the few-shot setting. We randomly select those examples for \texttt{GPT3.5FS}. For \texttt{GPT3.5Retr}, similar to semantic neighbors, we use \texttt{all-mpnet-base-v2} from SentenceBert \citep{reimers-gurevych-2019-sentence}, which performs best in semantic search to retrieve similar instances from the training set based on query $q$ in \S\ref{sec:retrieval}. The input length is limited to $2048$ tokens due to OpenAI API limits. We choose \texttt{gpt-4-0314} as our GPT4 model. Our input for GPT4 is similar to GPT3.5.

For each selected example from the training set with \textit{forward} relation, the template is \textit{``Consider the following context: $\mathcal{M}$ In that context, which $p$ can be used for $v$, and why? $\mathcal{T}$''}, where $\mathcal{M}$ is the background context, $p$ is the target node type, $v$ is the seed term, and $\mathcal{T}$ is the target idea sentence; for \textit{backward} relation, the template is \textit{``Consider the following context: $\mathcal{M}$ In that context, which $p$ do we use $v$, and why? $s$''}. For selected examples with additional retrieval inspirations, we concatenate the following additional template to the $\mathcal{M}$: \textit{``The retrieval results are: $i_1,\ldots,i_k$''}, where $i_1,\ldots,i_k$ are retrieved inspirations. For the final prompt, the template is similar to the above example template. However, the target sentence $\mathcal{T}$ will not be included. We ask the model to generate $10$ outputs. We will select the best output and skip the empty output.

\subsubsection{Fine Tuning}
\label{app:t5}
Given input without any inspirations, the input combines the prompt $\mathcal{P}$ and context $\mathcal{M}$ as shown in \S\ref{sec:retrieval} (i.e., \textit{$\mathcal{P}$ | context: $\mathcal{M}$}).  Given input with inspirations, the input is \textit{$\mathcal{P}$ | retrieve: $i_1,\ldots,i_k$ | context: $\mathcal{M}$}, with $i_1,\ldots,i_k$ as retrieved inspirations.
The input length is limited to $512$ tokens. For both tasks, we finetune our model based on \texttt{T5-large} with a learning rate of $6\times 10 ^{-6}$ and $\epsilon=1\times 10 ^{-6}$. The batch size is $8$ for each GPU. The maximum training epoch for all models is $10$ with $4$ patience.
During decoding, we use beam-search to generate results with a beam size of 5 and a repetition penalty of $1.5$. 

\paragraph{In-context Contrastive Augmentation}
We randomly select $2$ sentences that appeared in the input as in-context negatives. For example, in Figure~\ref{fig:overview}, the in-context negatives could be \textit{``knowledge acquisition is done by using Method''}, \textit{``this requires plms to integrate the information from all the sources in a lifelong manner .''}.

\subsubsection{Biochemical Case Study}
\label{app:biodetail}

Our Meditron-7b~\cite{chen2023meditron} and its variants are built based on the Huggingface framework \citep{wolf-etal-2020-transformers}.\footnote{\url{github.com/huggingface/transformers}} We use its \texttt{epfl-llm/meditron-7b} as the base model. We finetune those models with a learning rate of $2\times 10 ^{-6}$ and $\epsilon=5\times 10 ^{-8}$. The maximum training epoch for all models is $5$. All models are fine-tuned on 4 NVIDIA A100 80 GB GPUs with Fully Sharded Data Parallel.\footnote{\url{https://huggingface.co/docs/accelerate/usage_guides/fsdp}} The training time for each model is about 20 hours.

\begin{table*}[!htb]
\centering
\small
\begin{tabularx}{\linewidth}{>{\hsize=0.5\hsize}X>{\arraybackslash\hsize=1.5\hsize}X}
\toprule
\textbf{Type}&     \textbf{Content}   \\
\midrule
Seed Term Prompt                  & data augmentation is used for Task\\\hdashline
Context                     & data augmentation is an effective solution to data scarcity in low - resource scenarios. however, when applied to token-level tasks such as ner , data augmentation methods often suffer from token-label misalignment, which leads to unsatsifactory performance.\\\hdashline
Semantic Neighbors          & st and automatic speech recognition (asr), \underline{low-resource tagging tasks}, end-to-end speech translation, neural online chats response selection, neural machine translation, \underline{semi-supervised ner}, entity and context learning, semi-supervised setting, dependency parsing, low-resource machine translation, slot filling, dialog state tracking, visual question answering, visual question answering (vqa), low-resource neural machine translation\\\hdashline
KG Neighbors                &  nmt-based text normalization, task-oriented dialog systems, task-oriented dialogue system, low-resource languages (lrl), end-to-end speech translation, visual question answering (vqa), multiclass utterance classification, clinical semantic textual similarity, neural online chats response selection, context-aware neural machine translation\\\hdashline
Citation Neighbors          & Contextual Augmentation: Data Augmentation by Words with Paradigmatic Relations, \underline{An Analysis of Simple Data Augmentation for Named Entity Recognition}, Data Augmentation for Low-Resource Neural Machine Translation, \underline{DAGA: Data Augmentation with a Generation Approach for Low-resource Tagging Tasks}, EDA: Easy Data Augmentation Techniques for Boosting Performance on Text Classification Tasks\\\hdashline
Ground Truth  & ELM: Data Augmentation with Masked Entity Language Modeling for Low-Resource NER\\
\bottomrule
\end{tabularx}
\caption{Example (from \cite{zhou-etal-2022-melm}) of retrieved inspirations. Inspirations similar to ground truth are \underline{underlined}.\label{tab:neigh_s}}
\end{table*}

\subsection{The Scale of Retrieval Set}
We retrieve from a set of 59k papers with over 374k sentences in the NLP domain, the focus of our experiments. Our background KG built on the training set has more than 197k nodes and 261k relations. Moreover, we collect 87k paper titles from citation networks. This represents a large-scale and diverse domain; retrieving inspirations from this set is expected, in principle, to be more than enough for generating novel ideas. Indeed, NLP papers typically cite each other and build on each other as inspirations to create new ideas - which motivates our inspiration retrieval. 

\subsection{Automated Evaluation}
\label{sec:evalmetrics}
We use BERTScore~\cite{Zhang*2020BERTScore} with SciBERT checkpoint for both tasks. The hash of the checkpoint is \texttt{allenai/scibert\_scivocab\_uncased\_L8} \texttt{\_no-idf\_version=0.3.12(hug\_trans=4.19.2)}. The automated evaluation results are in Table~\ref{tab:sp_ncs}.
\begin{figure*}[htb]
\centering
\includegraphics[width=\linewidth]{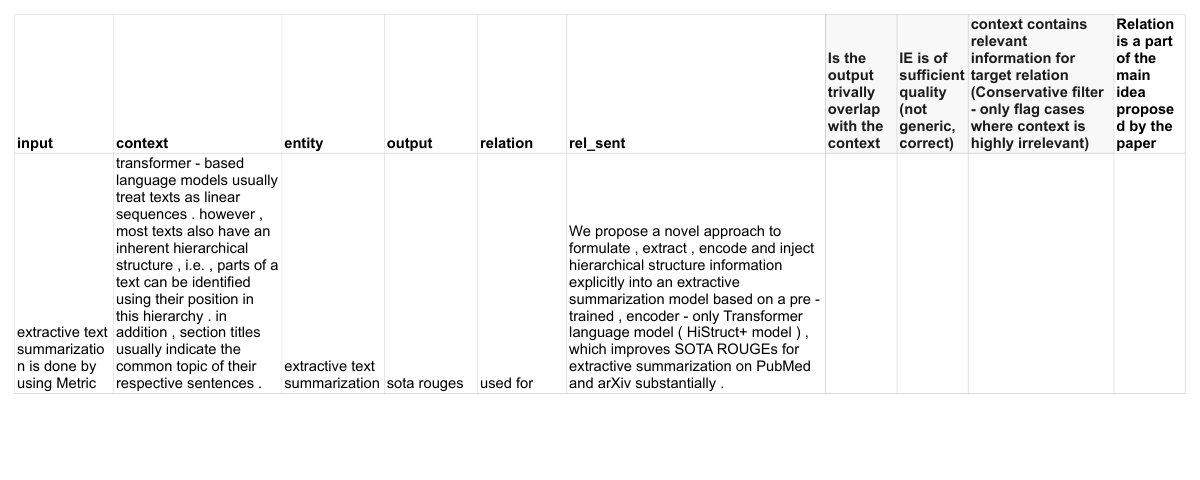}
\caption{\label{fig:gi}Gold subset annotation interface
}
\end{figure*}
\begin{table}[!htb]
\centering
\small
\begin{tabularx}{\linewidth}{>{\hsize=1.4\hsize}X>{\centering\arraybackslash\hsize=0.8\hsize}X>{\centering\arraybackslash\hsize=1\hsize}X>{\centering\arraybackslash\hsize=0.8\hsize}X>{\centering\arraybackslash\hsize=1\hsize}X} 
\toprule
\textbf{Subset}  &\multicolumn{2}{c}{\textbf{Challenging}}&\multicolumn{2}{c}{\textbf{Gold}}\\ 
\midrule
\textbf{Model}  &\textbf{R-L$\uparrow$}&\textbf{BERT$\uparrow$}&\textbf{R-L$\uparrow$}&\textbf{BERT$\uparrow$}\\ 
\midrule

\texttt{GPT4ZS}  & 0.120     & 0.581      & 0.130    & 0.583 \\
\texttt{GPT4FS}  & 0.143     & 0.618      & 0.151    & 0.624 \\

\texttt{T5}                & 0.223     & \textbf{0.672}$^\dag$  &  0.246   & 0.685\\ 
\hdashline

\texttt{GPT4FS+SN}          & 0.144& 0.620      & 0.149    & 0.627 \\
\texttt{GPT4FS+KG}          & 0.143& 0.619      & 0.152    & 0.626 \\
\texttt{GPT4FS+CT}          & 0.144& 0.617      & 0.149    & 0.622 \\

\hdashline
\texttt{T5+CL}           & 0.225$^\dag$     & 0.671$^\dag$  &  0.251$^\dag$   & 0.686$^\dag$\\ 
\texttt{T5+SN+CL}      & \textbf{0.228}$^\dag$     & 0.671$^\dag$  &  \textbf{0.258}$^\dag$   & 0.686$^\dag$  \\ 
\texttt{T5+KG+CL}      & 0.223$^\dag$     & 0.669  &  0.248   & 0.681$^\dag$ \\ 
\texttt{T5+CT+CL}     & 0.225$^\dag$     & 0.671$^\dag$ &  0.250$^\dag$   & \textbf{0.686}$^\dag$  \\ 
\bottomrule
\end{tabularx}

\caption{Automatic evaluation results for the challenging and gold subsets. \textit{CL} is a model with in-context contrastive augmentation. \textit{SN} is a model with semantic inspirations. \textit{KG} is a model with KG inspirations. \textit{CT} is a model with citation inspirations. \textit{R-L} denotes ROUGE-L. \textit{BERT} denotes BERTscore with SciBERT as its encoder. $^\dag$ indicates that differences between models are not statistically significant ($p \leq 0.05$) when compared to each other but are still significant when compared to the other models based on t-test. 
\label{tab:sp_ncs} }
\end{table}

\begin{table*}[!htb]
\centering
\small
\begin{tabularx}{\linewidth}{>{\hsize=0.4\hsize}X>{\arraybackslash\hsize=1.6\hsize}X}
\toprule
\textbf{Type}&\textbf{Content}    \\
\midrule
Seed Term Prompt & Irish language learning is done by using Method\\\hdashline
Context & Irish is a minority language which means that l2 learners have limited opportunities for exposure to the language, and additionally, there are also limited (digital) learning resources available.\\
\bottomrule
\end{tabularx}
\caption{Input for sample human annotation results\label{tab:human_input}
}
\end{table*}
\begin{table*}[!htb]
\centering
\small
\begin{tabularx}{\linewidth}{>{\hsize=0.4\hsize}X>{\arraybackslash\hsize=2.3\hsize}X>{\centering\arraybackslash\hsize=0.3\hsize}X}
\toprule
\textbf{Model}&\textbf{Output}    &   \textbf{Label}  \\
\midrule
\texttt{GPT3.5FS}&The use of Social Media in \textbf{Irish Language Learning}: A Case Study& \textit{helpful}\\\hdashline
\texttt{GPT3.5Retr}& One method that could be used for \textbf{Irish language learning} is computer assisted language learning (CALL)& \textit{unhelpful}\\\hdashline
\texttt{GPT3.5FS+CT}&In that context, the most appropriate task for precision grammar would be the development of a part-of-speech tagger for Irish using finite-state morphology and constraint grammar disambiguation.& \textit{unhelpful} \\\hdashline
\texttt{GPT3.5FS+KG}&Cló: An \textbf{Irish Language Learning} Platform - Leveraging Digital Resources and Human Expertise& \textit{unhelpful}\\\hdashline
\texttt{GPT4FS}&We propose using a technology-enhanced approach, such as computer-assisted language learning (CALL) or natural language processing (NLP) techniques, to create interactive and engaging \textbf{Irish language learning} resources, as these methods can provide personalized learning experiences and enhance learners' exposure to the language despite limited resources available.& \textit{helpful}\\\hdashline
\texttt{GPT4FS+KG}&Exploiting Social Media for Irish Language Learning: An Analysis of Twitter Data. In this context, we use social media data, particularly from Twitter, as a method for \textbf{Irish language learning}, because it provides a rich source of authentic and diverse language examples that can be used to enhance learning opportunities for L2 learners in a minority language setting.& \textit{helpful}\\\hdashline
\texttt{T5}&In this paper, we present a novel machine translation approach to \textbf{Irish language learning}.&\textit{unhelpful} \\\hdashline
{\texttt{T5+SN+CL}}&
A conversational virtual environment for \textbf{Irish language learning}& \textit{helpful}\\
\bottomrule
\end{tabularx}
\caption{Sample human labels for idea-sentence generation results for seed term prompt \textit{``\textbf{Irish language learning} is done by using Method''}, with context  \textit{``irish is a minority language ... limited digital learning resources available''}. \label{tab:human_anno}}
\end{table*}

\begin{table*}[!htb]
\centering
\small
\begin{tabularx}{\linewidth}{>{\hsize=0.4\hsize}X>{\arraybackslash\hsize=1.6\hsize}X}
\toprule
\textbf{Type}&\textbf{Content}    \\
\midrule
Seed Term & ARO10\\\hdashline
Context & Huangjiu is known for its unique aroma, primarily attributed to its high concentration of b-phenylethanol (ranging from 40 to 130 mg/L).\\\hdashline
Meditron & We then selected six key genes involved in the phenylalanine metabolism pathway and predicted that their product flux may affect the synthesis of b-phenylethanol.
\\\hdashline
Meditron + CT& 
We found that the key amino acid residue that controls the activity of \textbf{Aro10p} was not conserved in wine yeast strains, which may explain the lower b-phenylethanol production in wine fermentation compared with that in Chinese huangjiu.
\\\hdashline
Meditron + KG& 
Both target genes, SSA1 and \textbf{ARO10}, were deleted using the CRISPR-Cas9 genome editing system.
\\\hdashline
Meditron + SN& 
Herein, we report that the key barrier for b-phenylethanol production in Huangjiu is \textbf{ARO10}, the only bi-functional amino acid decarboxylase in Saccharomyces cerevisiae.
\\
\bottomrule
\end{tabularx}
\caption{Input and idea-sentence generation results for seed gene ``\textbf{\textit{ARO10}}'' in the biochemical domain\label{tab:bio_example}.
}
\end{table*}
\begin{figure*}[htb]
\centering
\includegraphics[width=\linewidth]{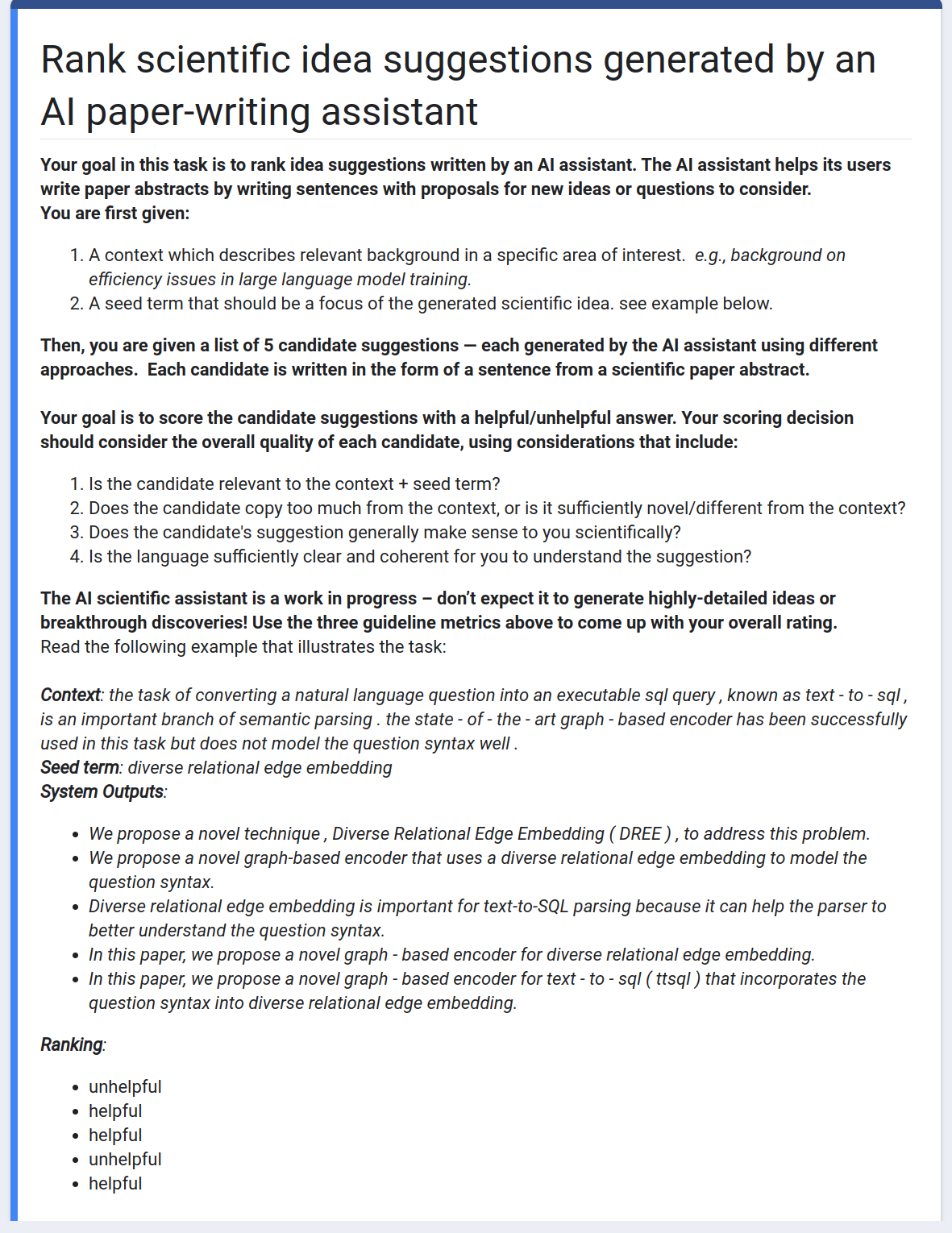}
\caption{\label{fig:hi}Human evaluation instructions
}
\end{figure*}
\begin{figure*}[htb]
\centering
\includegraphics[width=0.8\linewidth]{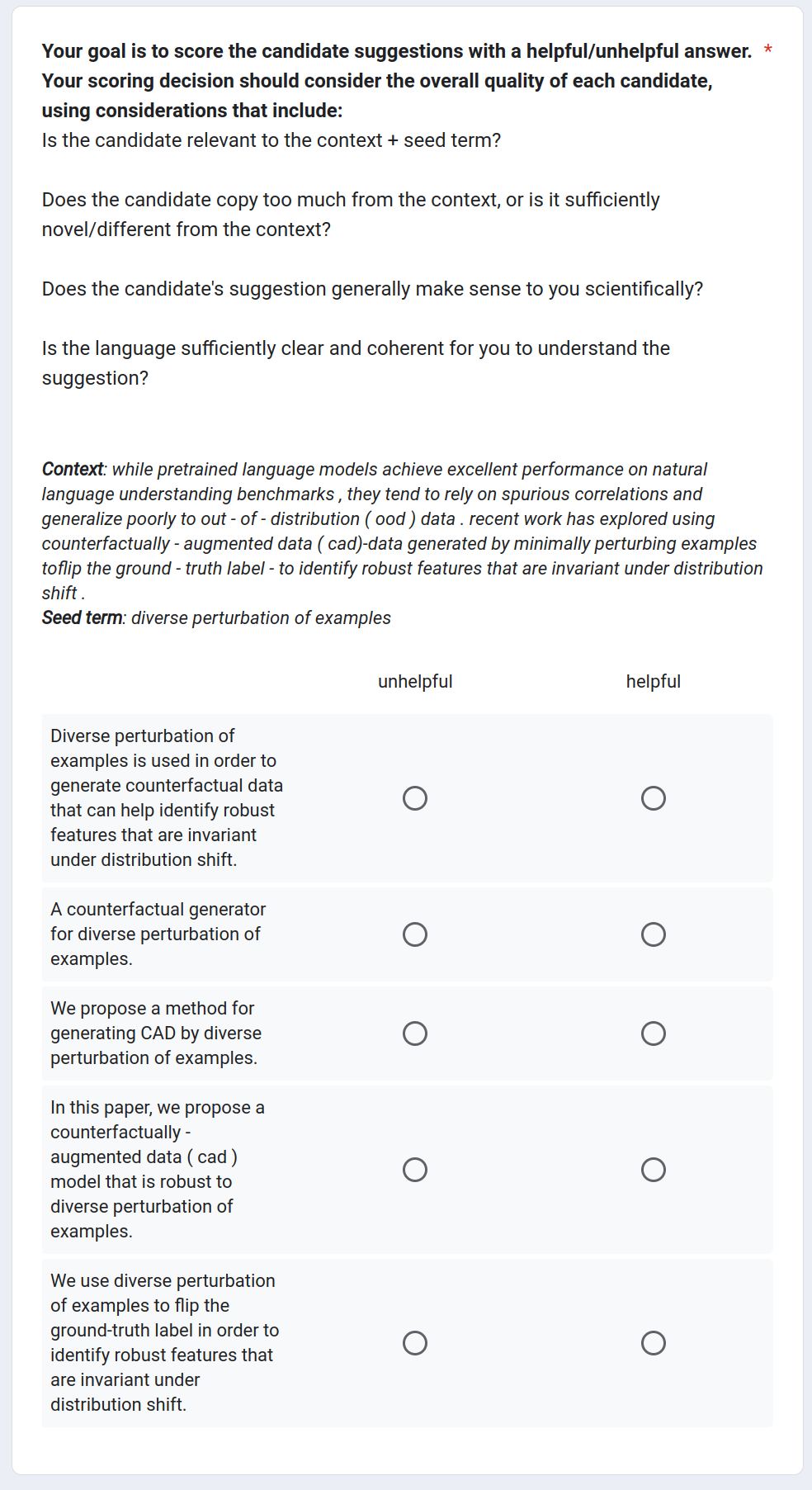}
\caption{\label{fig:hexp}Human evaluation example for  \texttt{GPT3.5Rnd}, \texttt{GPT3.5Retr}, \texttt{GPT3.5Rnd+CT}, \texttt{T5}, and {\texttt{T5+SN+CL}}
}
\end{figure*}

\begin{figure*}[htb]
\centering
\includegraphics[width=0.6\linewidth]{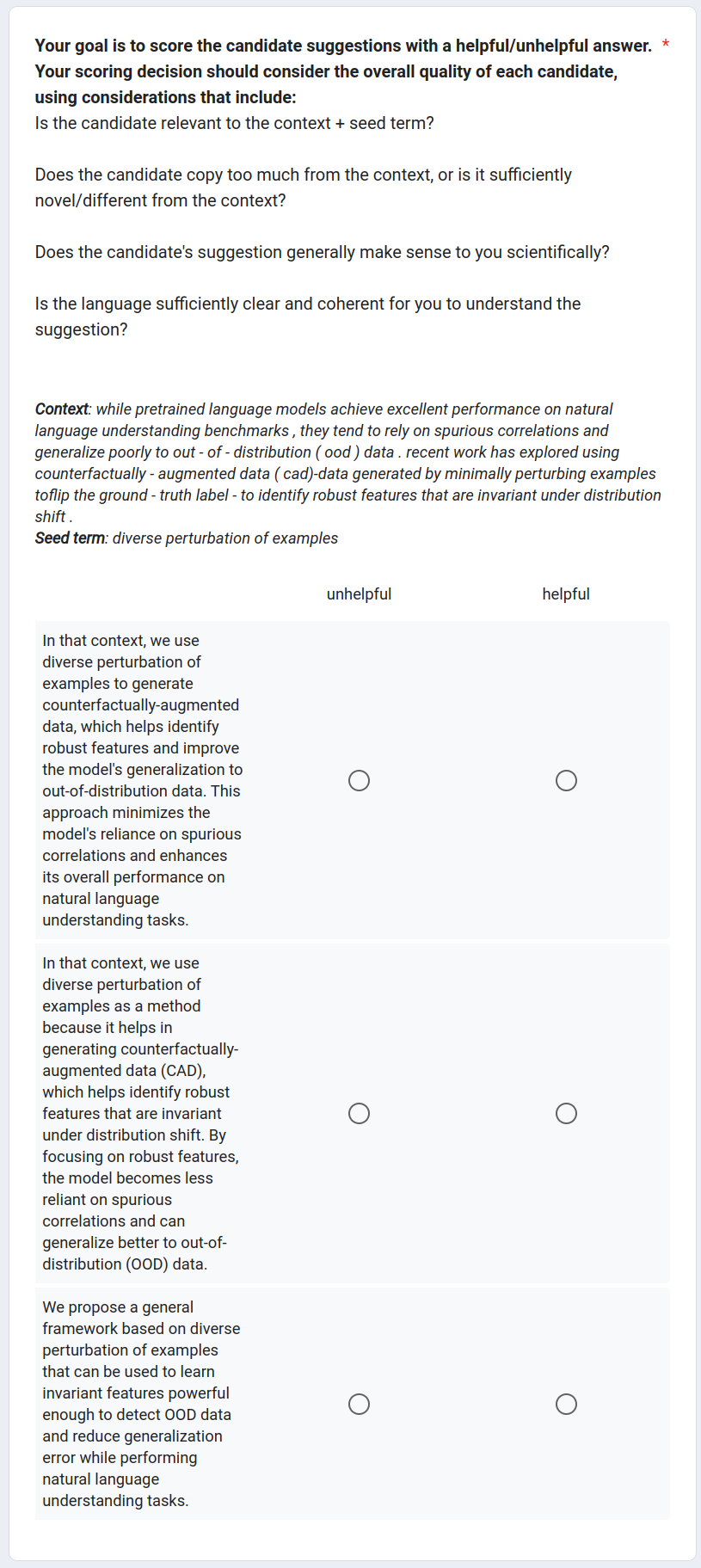}
\caption{\label{fig:hexp2}Human evaluation example for \texttt{GPT3.5Rnd+KG}, \texttt{GPT4Rnd}, and \texttt{GPT4Rnd+KG}
}
\end{figure*}

\section{Human Annotation and Evaluation Details}

\label{app:he_ap}

\paragraph{Gold Dataset Annotation Details}
The gold dataset annotation interface is in Figure~\ref{fig:gi}.  The quality of the instances in the test set is judged given three criteria: (1) whether the ground truth sentence trivially overlaps with background context; (2) whether background context contains relevant information for the target relation; (3) whether the target relation (from which the seed term is taken) is a salient aspect of the idea proposed in the target paper.

\paragraph{Study I}
The instructions for human evaluation can be found in Figure~\ref{fig:hi}, while an example of the human evaluation interface is provided in Figure~\ref{fig:hexp} and ~\ref{fig:hexp2}. Human annotators are required to evaluate each system output based on the following criteria: (1) \textit{Is the candidate relevant to the context + seed term?} (2) \textit{Does the candidate copy too much from the context, or is it sufficiently novel/different from the context?} (3) \textit{Does the candidate's suggestion generally make sense to you scientifically?} (4) \textit{Is the language sufficiently clear and coherent to understand the suggestion?} The input for sample human annotation is in Table~\ref{tab:human_input} and the human labels are in Table~\ref{tab:human_anno}. The human annotation agreement is in Table~\ref{tab:alpha}.

\paragraph{Study III}
We ask the following questions to human annotators to evaluate the quality of regeneration results: (1) \textit{Is the regenerated idea substantially different from the original? } (2) \textit{Is the regenerated idea more novel and creative than the original idea?} (3) \textit{Does the second iteration increase novelty?} The human annotation agreement is in Table~\ref{tab:alpha3}.

\begin{table}[!htb]
\centering
\small
\begin{tabularx}{\linewidth}{>{\arraybackslash\hsize=2.5\hsize}X>{\centering\arraybackslash\hsize=0.7\hsize}X>{\centering\arraybackslash\hsize=0.7\hsize}X>{\centering\arraybackslash\hsize=0.7\hsize}X>{\centering\arraybackslash\hsize=0.7\hsize}X>{\centering\arraybackslash\hsize=0.7\hsize}X}
\toprule
\textbf{Annotator Pair}&\texttt{1-2}& \texttt{1-3}& \texttt{1-4}& \texttt{1-5}& \texttt{1-6}  \\ 
\midrule
\textbf{Agreement \%}&68.8& 75.0& 56.2& 43.8& 75.0\\
\bottomrule
\end{tabularx}
\caption{ Percent (\%) of same labels from overlapped 10 human evaluation instances on each pair of annotators for Study I. 
\label{tab:alpha}
}
\end{table}

\begin{table}[!htb]
\centering
\small
\begin{tabularx}{\linewidth}{>{\arraybackslash\hsize=1.8\hsize}X>{\centering\arraybackslash\hsize=0.8\hsize}X>{\centering\arraybackslash\hsize=0.8\hsize}X>{\centering\arraybackslash\hsize=0.8\hsize}X>{\centering\arraybackslash\hsize=0.8\hsize}X}
\toprule
\textbf{Annotator Pair}&\texttt{1-2}& \texttt{1-3} & \texttt{1-4}& \texttt{1-5}  \\ 
\midrule
\textbf{Agreement \%}&92.5 & 93.3& 87.5& 90.0\\
\bottomrule
\end{tabularx}
\caption{ Percent (\%) of same labels from overlapped 20 human evaluation instances on each pair of annotators for Study III. (1-3) has 60 shared questions. The rest of the pairs each share 40 questions.
\label{tab:alpha3}
}
\end{table}

\section{Scientific Artifacts}
We list the licenses of the scientific artifacts used in this paper: Semantic Scholar Academic Graph API (API license agreement\footnote{\url{api.semanticscholar.org/license/}}), Huggingface Transformers (Apache License 2.0), SBERT (Apache-2.0 license), 
BERTScore (MIT license), 
Meditron-7b (Llama2), Entrez Programming Utilities API (Copyright\footnote{\url{www.ncbi.nlm.nih.gov/books/about/copyright/}}), PubTator 3 (Data use policy\footnote{\url{www.ncbi.nlm.nih.gov/home/about/policies/}}), and OpenAI (Terms of use\footnote{\url{openai.com/policies/terms-of-use}}).

\section{Ethical Consideration}
The \textsc{SciMON} task and corresponding models we have designed in this paper are limited to the natural language processing (NLP) and biochemical domain, and might not apply to other scenarios.

\subsection{Usage Requirement}
This paper aims to provide investigative leads for a scientific domain, specifically natural language processing. The final results are not intended to be used without human review. Accordingly, domain experts might use this tool as a research writing assistant to develop ideas. However, our system does not do any fact-checking with external knowledge. In addition, we train our models on the ACL anthology and PubMed papers written in English, which might alienate readers who have been historically underrepresented in the NLP/biochemical domains.

\subsection{Data Collection}
We collect 67,408 ACL Anthology papers from 1952 to 2022 using Semantic Scholar Academic Graph API, under API license agreement\footnote{\url{https://api.semanticscholar.org/license/}}. We ensure our data collection procedure follows the Terms of Use at \url{https://allenai.org/terms}. According to the agreement, our dataset can only be used for non-commercial purposes. As mentioned in \S\ref{sec:exp}, we perform the human evaluation. All annotators involved in human evaluation are voluntary participants with a fair wage. We further collect 5,708 PubMed papers from 1988 to 2024 using Entrez Programming Utilities API\footnote{\url{www.ncbi.nlm.nih.gov/books/NBK25501/}}. We follow their data usage guidelines\footnote{\url{www.ncbi.nlm.nih.gov/books/about/copyright/}}.

\end{document}